\documentclass[letterpaper]{article} % DO NOT CHANGE THIS
\usepackage{aaai2026}
\usepackage{times}  % DO NOT CHANGE THIS
\usepackage{helvet}  % DO NOT CHANGE THIS
\usepackage{courier}  % DO NOT CHANGE THIS
\usepackage[hyphens]{url}  % DO NOT CHANGE THIS
\usepackage{graphicx} % DO NOT CHANGE THIS
\usepackage{natbib}  % DO NOT CHANGE THIS AND DO NOT ADD ANY OPTIONS TO IT
\usepackage{caption} % DO NOT CHANGE THIS AND DO NOT ADD ANY OPTIONS TO IT
\usepackage{algorithm}
\usepackage{newfloat}
\usepackage{listings}
\DeclareCaptionStyle{ruled}{labelfont=normalfont,labelsep=colon,strut=off} % DO NOT CHANGE THIS
\floatstyle{ruled}
\newfloat{listing}{tb}{lst}{}
\floatname{listing}{Listing}
\usepackage{amsmath}   % For advanced math environments and commands
\usepackage{amssymb}   % For additional math symbols (like \mathbb)

\usepackage{booktabs}  % For professional-quality rules in tables (e.g., \toprule)
\usepackage{multirow}  % To span table cells across multiple rows
\usepackage{tabularx}  % For tables with columns that automatically adjust width
\usepackage{array}

\usepackage{algpseudocode} 

\usepackage{subfig}

\usepackage{siunitx}      % For professionally typesetting numbers and units (e.g., \SI{10}{\kilo\gram})
\usepackage[autostyle]{csquotes} % For context-sensitive quotation marks (e.g., \enquote{})
\usepackage[capitalise]{cleveref}     % For smart cross-referencing with \cref{}
\usepackage{seqsplit}

\DeclareUnicodeCharacter{0301}{\'{e}} % Handles a specific Unicode character for accented 'e'
\crefname{subsubsection}{subsubsection}{subsubsections}
\Crefname{subsubsection}{Subsubsection}{Subsubsections}

\title{BSAT: B-Spline Adaptive Tokenizer\\for Long-Term Time Series Forecasting}

\title{BSAT: B-Spline Adaptive Tokenizer for Long-Term Time Series Forecasting}
\author {
    Maximilian Reinwardt\textsuperscript{\rm 1},
    Michael Eichelbeck\textsuperscript{\rm 1},
    Matthias Althoff \textsuperscript{\rm 1}
}
\affiliations {
    \textsuperscript{\rm 1}
Technical University of Munich, Munich, Germany

\{max.reinwardt, michael.eichelbeck, althoff\}@tum.de\\
}

\begin{document}

\maketitle

\begin{abstract}
Long-term time series forecasting using transformers is hampered by the quadratic complexity of self-attention and the rigidity of uniform patching, which may be misaligned with the data's semantic structure. In this paper, we introduce the \textit{B-Spline Adaptive Tokenizer (BSAT)}, a novel, parameter-free method that adaptively segments a time series by fitting it with B-splines. BSAT algorithmically places tokens in high-curvature regions and represents each variable-length basis function as a fixed-size token, composed of its coefficient and position. 
Further, we propose a hybrid positional encoding that combines a additive learnable positional encoding with Rotary Positional Embedding featuring a layer-wise learnable base: \textit{L-RoPE}. This allows each layer to attend to different temporal dependencies. Our experiments on several public benchmarks show that our model is competitive with strong performance at high compression rates. This makes it particularly well-suited for use cases with strong memory constraints.\end{abstract}


\section{Introduction}

Long-term time series forecasting plays a crucial role across many sectors \cite{cirstea_towards_2022, perveen_handling_2020, pan_magicscaler_2023}. 
Deep learning models, particularly transformers \cite{vaswani_attention_2017}, excel at capturing long-range dependencies \cite{lara-benitez_evaluation_2021, li_deep_2024}.
However, as their point-wise self-attention scales quadratically in both computation and memory, they become impractical for very long sequences. 
%Early Time Series Transformers (TSTs) like Informer \cite{zhou_informer_2021}, Autoformer \cite{wu_autoformer_2022} and FEDformer \cite{zhou_fedformer_2022} primarily focused on mitigating this complexity architecturally, but proved unable to consistently outperform simple linear models like DLinear \cite{zeng_are_2023}.
PatchTST \cite{nie_time_2023} improved efficiency by segmenting sequences into overlapping fixed-length sub-sequences, \textit{patches}. %, while making full use of the self-attention architecture. 
Conversely, this uniform segmentation can inadvertently split meaningful patterns or waste computational capacity on less informative regions \cite{huang_hdmixer_2024, liu_rethinking_2025, cao_mspatch_2025, pagnoni_byte_2024}.
To overcome the rigidity of patching, we explore \textit{adaptive tokenization}, where tokens are dynamically aligned with the structure of the series. This introduces new challenges:
(i) defining semantically meaningful tokens;
(ii) embedding tokens of varying lengths; and
(iii) accurately encoding non-uniform positions.

In this paper, we address these challenges through two major contributions:
(1) We introduce the \textit{B-Spline Adaptive Tokenizer (BSAT)}, a novel tokenization strategy that adaptively segments time series using a B-spline curve. BSAT algorithmically places more tokens where the series exhibits high complexity. Each token encodes one basis function and is composed of two scalars: coefficient and position.  

(2) To embed the positions of these variable-sized and overlapping tokens, we introduce a hybrid positional encoding strategy, applying a learnable positional encoding and Rotary Positional Embedding (RoPE) \cite{su_roformer_2023}: Hybrid Additive Rotary Positional Encoding. Additionally, we propose \textit{L-RoPE}, a layer-wise learnable RoPE frequency base, enabling the model to capture dataset-specific temporal patterns. 

Our experiments show that BSAT achieves strong performance against common baseline models and notably, performs particularly well on low token budgets. 
We argue that, the layer-wise learnable RoPE base acts as a multi-resolution attention mechanism.

\section{Related Work}
\paragraph{Patching in Time Series Transformers}
Segmentation of large input sequences has proven successful across domains \cite{devlin_bert_2019, pagnoni_byte_2024, dosovitskiy_image_2021}. In the long-term time series forecasting domain, PatchTST \cite{nie_time_2023} segments time series into uniform-size patches, treating them as input tokens. This improved both performance and efficiency by capturing local semantic information and shrinking sequence length \cite{wang_deep_2024}. %Since its inception patching has been shown to be successful with alternative backbones like multilayer perceptrons \cite{ekambaram_tsmixer_2023}, CNNs \cite{gong_patchmixer_2024}, and in large scale pre-trained models like TimesFM \cite{das_decoder-only_2024}.

However, uniform patches create challenges \cite{pagnoni_byte_2024, huang_hdmixer_2024, liu_rethinking_2025, cao_mspatch_2025}:
(1) They may segment semantic structures, peaks, and periodic patterns or fail to split short structures, causing information loss.
(2) They produce uneven information density per token. %, as not all sub-sequences hold equal information content or equally critical dependencies. 
As models assign an equal amount of compute to each token, a substantial share is spent inefficiently.
%PatchTST aim to address this by overlapping patches by 50\%, to maximize local information. Conversely this also means every point is processed twice.
To address this, some Time Series Transformers (TSTs) leverage multiple fixed patch sizes in parallel \cite{chen_pathformer_2024, cao_mspatch_2025, du_multiresformer_2024}.  
Studies in MPLs have introduced adjustable patches \cite{huang_hdmixer_2024, liu_rethinking_2025}.
In large, pre-trained time series models, adaptive tokenization via a learnable selection from a fixed set of patch sizes \cite{kamarthi_large_2024}, and variable motif-based segmentation \cite{gotz_byte_2025} have been proposed.
These approaches have limitations: Parallel patch sizes increase compute costs, the flexibility of adjustable patches is limited, and they are incompatible with TSTs. Pre-trained tokenizers transfer poorly across domains and require substantial data for training. Further embedding variate patch sizes is challenging. Existing Options \cite{nie_time_2023, kamarthi_large_2024, du_multiresformer_2024, gotz_byte_2025}, all result in information loss, wasted compute or added architectural complexity.

\paragraph{B-Splines}
B-splines are piecewise polynomial functions for constructing flexible curves with local control and adjustable smoothness. 
This is achieved by representing them as the linear combination of basis functions, weighted by coefficients. 
%These properties have led to widespread applications \cite{chaudhuri_b-splines_2021, hasan_b-spline_2024, liu_b-spline-based_2019, eilers_flexible_1996, ramsay_functional_2005, hastie_generalized_1986, li_time_2024}.
Some deep learning approaches have integrated splines into recurrent neural networks and multilayer perceptrons \cite{hajiabotorabi_improving_2019, kong_nonlinear_2018, bilos_irregularly-sampled_2022, gasthaus_probabilistic_2019}. 
Notably, BasisFormer, learns global basis vectors, and predicts based on similarity to global patterns \cite{ni_basisformer_2024}. 
To the best of our knowledge, a TST with an input token representing B-spline basis functions, enabling adaptive token lengths, has not been proposed before. 

\paragraph{Non-Integer Relative Positional Encoding}
The permutation-invariant self-attention mechanism necessitates an explicit encoding of sequence order and positions \cite{huang_improve_2020}.  Many models use fixed or learned absolute positional embeddings \cite{vaswani_attention_2017, devlin_bert_2019, nie_time_2023}. For time series with long, complex dependencies, relative positional embeddings and hybrid positional embeddings \cite{huang_improve_2020, ke_rethinking_2021, liutkus_relative_2021, zhang_exploring_2024} have improved performance \cite{irani_positional_2025}.RoPE \cite{su_roformer_2023}, a relative positional encoding method has been widely adopted in natural language processing \cite{touvron_llama_2023} and applied to irregular time series \cite{zivanovic_rotary_2025}. RoPE Base modification can serve to control attention decay, biasing attention toward short- or long-range patterns \cite{men_base_2024}. Some studies have explored learning the RoPE base pair-wise \cite{zhang_elastst_2024, heo_rotary_2024}, to better adapt attention patterns to the data. 
Motivated by this, we propose two modifications: a layer-wise learnable base, allowing each layer to attend to distinct temporal patterns and a hybrid positional encoding that applies both additive and rotary embeddings.

\section{Preliminaries}

\subsection{B-Splines}
B-splines offer a principled framework for adaptive time series segmentation, through their desirable mathematical properties (partition of unity, continuity, and provably optimal local approximations), computational advantages (linear basis structure and compact support enabling natural tokenization), and well-established signal-based knot placement algorithms from approximation theory.
B-splines are defined as $n = k-(p+1)$ piecewise polynomial functions defined over a non-decreasing knot sequence $\boldsymbol{\tau} = (\tau_0, \dots, \tau_{k-1})$ with $k$ knots, parameterized by their degree $p$ and clamped by $p+1$ boundary knots.
For $\xi\in[\tau_p,\tau_{n}]\subset\mathbb{R}$, the $i$-th B-spline basis function $N_{i,p}(\xi)$ is defined recursively by the Cox--de Boor relation~\cite[Ch. 9]{de_boor_practical_2001}:
\begin{align}
N_{i,p}(\xi) =\;& \frac{\xi - \tau_i}{\tau_{i+p} - \tau_i} N_{i,p-1}(\xi) \notag \\
& + \frac{\tau_{i+p+1} - \xi}{\tau_{i+p+1} - \tau_{i+1}} N_{i+1,p-1}(\xi).
\end{align}
Each basis function has local support over the interval $[\tau_i, \tau_{i+p+1})$; therefore, it only affects a limited region of the curve. Given a set of coefficients $\{c_i\}_{i=0}^{n-1}$, a B-spline curve is then defined as
\begin{equation}
C(\xi) = \sum_{i=0}^{n-1} c_i N_{i,p}(\xi).
\end{equation}
To construct a B-spline curve, first the knot vector must be determined. %Then the coefficients can be calculated. 
For knot placement we adapt a derivative based approach from Yeh et al. \cite{yeh_fast_2020}. 
There, for a windowed signal $y$, a per-sample feature function $f$ is defined as
\[
f^{\text{Yeh}}_i = \bigl|y^{(p)}(\xi_i)\bigr|^{2/p}.
\]
The parameter domain $[\xi_{\min}, \xi_{\max}]$ is then discretized into $u_0, \dots, u_m$.
This feature function is then integrated over each sub-interval $[u_i, u_{i+1}]$ using the trapezoidal rule, yielding interval masses
\begin{equation}
w_i = \frac{ u_{i+1} - u_i}{2} \left[ f(u_i) + f(u_{i+1}) \right].
\end{equation}
To control knot concentration, \cite{yeh_fast_2020} introduce clip factor $g=1$ and clip $w_i$ to $\tilde{w}_i = \min(w_i,\, g\Delta F)$, where $\Delta F=\int_{\xi_{\min}}^{\xi_{\max}} f\,\frac{d\xi}{m}$ is the mean interval mass required for $k_{int} = k-2(p+1)$ interior knots.
A cumulative distribution function of $\tilde{w}$ is inverted at uniform quantiles to place $k_{int}$ interior knots. 
A pseudo code implementation is available in \cref{app:bsat}.

\subsection{Transformers}
Given a token sequence $\mathbf{X}\!\in\!\mathbb{R}^{n\times d_{\text{model}}}$, a single scaled dot‑product attention layer as introduced in~\cite{vaswani_attention_2017},
computes
\begin{equation}
\text{Attn}(\mathbf{Q},\mathbf{K},\mathbf{V})
    = \mathrm{softmax}\!\bigl(\mathbf{Q}\mathbf{K}^{\top}/\sqrt{d_{head}}\bigr)\mathbf{V},
\label{eq:attn}
\end{equation}
where queries, keys and values are linear projections:
$\mathbf{Q}=\mathbf{X}W^{Q},\;\mathbf{K}=\mathbf{X}W^{K},\;\mathbf{V}=\mathbf{X}W^{V}$
with $\;(W^{\{\cdot\}}\!\in\!\mathbb{R}^{d_{model}\times d_{head}})$.
To enrich the representation space, Transformer blocks split the model dimension into $h$ independent heads.
Each head operates on $d_{head}\!=\!d_{\text{model}}/h$ features using its own set of projection matrices.
The per-head attention for head $j$ is
\[
\text{head}_j=\mathrm{softmax}\!\bigl(\mathbf{Q}_j\mathbf{K}_j^{\top}/\sqrt{d_{head}}\bigr)\mathbf{V}_j,
\]
composing the multi-head attention
\[
\text{MHA}(\mathbf{X})=\text{Concat}(\text{head}_1,\ldots,\text{head}_h)\,W^{O},
\]
with $W^{O}\in\mathbb{R}^{(h\,d_{head})\times d_{model}}$.

\subsection{Rotary Positional Embeddings}
Because the attention mechanism in \eqref{eq:attn} is permutation‑invariant, positional information is typically injected additively via sinusoids or learned embeddings, or multiplicatively via RoPE \cite{su_roformer_2023}.  
It encodes relative position information into the attention computation through head-wise geometric transformations. 
Given query and key vectors $\mathbf{q}, \mathbf{k} \in \mathbb{R}^{d_{head}}$ with even head dimension $d_{head}$, RoPE divides each vector into $d_{head}/2$ two-dimensional sub-vector pairs:
\begin{equation}
\mathbf{q} = [\mathbf{q}^{(1)}, \mathbf{q}^{(2)}, \ldots, \mathbf{q}^{(d_{head}/2)}], \quad \mathbf{q}^{(i)} \in \mathbb{R}^2.
\end{equation}

For each pair, RoPE applies a rotation based on the position of the token.  
The rotation angle for integer position $a$ and dimension pair $i$ is determined by
\begin{equation}
\theta_i = a \cdot f_i, \quad \text{where} \quad f_i = \theta_{\text{base}}^{-2(i-1)/d_{head}}.
\end{equation}
Here, $\theta_{\text{base}}$ (typically 10,000) controls the frequency spectrum, with $f_1 = 1$ (fastest rotation: every $2  \pi\approx~6.28$ steps) and $f_{d_{head}/2} \approx 1/\theta_{\text{base}}$ (slowest rotation).
All layers and heads reuse the same frequency spectrum $f_i$.
This same approach can be applied to real, non-integer positions \cite{zivanovic_rotary_2025}.

\section{Methodology}
\subsection{Problem Statement}
Let $x_{1:T} \in \mathbb{R}^{T}$ be a univariate time series sampled at a fixed interval $\delta$.
At a time step $t \in \mathbb{N}$, we observe the lookback window $\mathbf{x}_{t} = (x_{t-L+1}, \dots, x_t) \in \mathbb{R}^L$.
Our objective is to predict the future values $\mathbf{y}_{t} = (x_{t+1}, \dots, x_{t+H}) \in \mathbb{R}^H$ using a model that operates on a compressed representation of $\mathbf{x}_{t}$.
Specifically, let $h_\phi: \mathbb{R}^L \to \mathbb{R}^{d \times n}$ denote a tokenization function that maps the lookback window to $n \ll L$ tokens.
The trainable prediction model $f_\theta$ then operates on this representation:
$\hat{\mathbf{y}}_{t} = f_\theta\big( h_\phi(\mathbf{x}_{t}) \big).$
We seek to find $h_\phi$ that minimizes both the compression ratio $n/L$ and the prediction error $\|\hat{\mathbf{y}}_{t} - \mathbf{y}_{t}\|$.
%All variables are listed in Appendix~\cref{app:var}.

\subsection{BSAT: B-Spline Adaptive Tokenizer}
\begin{figure}[t]
  \centering
  \includegraphics[width=0.92\columnwidth]{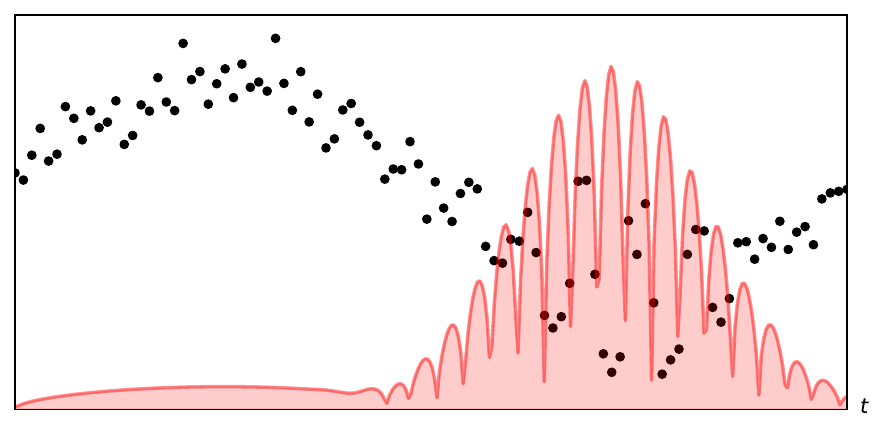}\\
  \includegraphics[width=0.92\columnwidth]{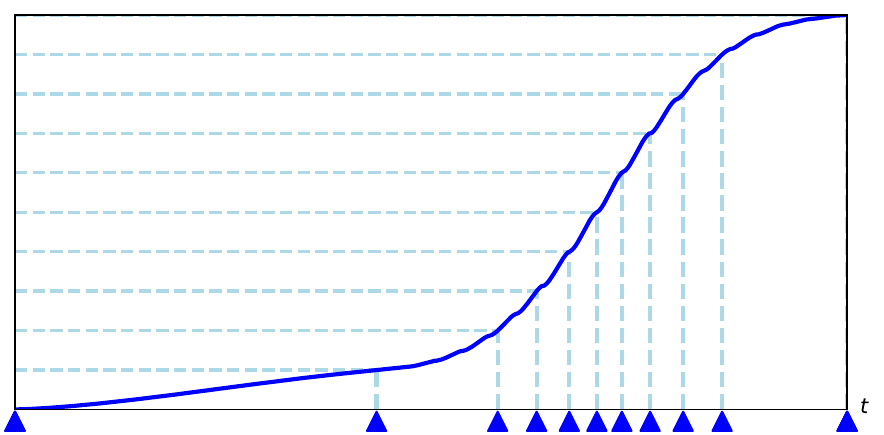}\\
  \includegraphics[width=0.92\columnwidth]{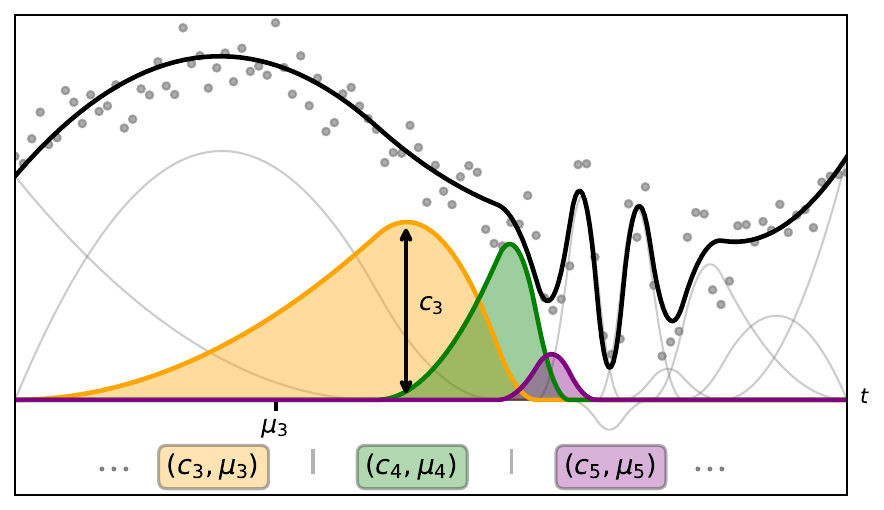}
  \caption{Visualization of BSAT mapping 100 points into 12 tokens. Knot placement is densest in high-curvature segments. Top: Time series observations (black) and the resulting feature function (red). Middle: cumulative distribution function with quantiles and the resulting knot positions (blue triangles). Bottom: Observations, the final fitted B-spline curve, the basis functions that compose it, and the tokens that result from the basis functions. Each token is constructed of one coefficient $c_i$ and one center $\mu_i$.}
  \label{fig:pipeline_stacked}
\end{figure}

We propose a novel tokenization strategy to mitigate the quadratic complexity of self-attention and optimize computational allocation.
BSAT segments time series into adaptively sized, overlapping tokens.
This is achieved by algorithmically assigning more, smaller tokens to regions exhibiting higher curvature. This increased token density enhances accuracy and allocates more computing resources to complex areas. This is visualized for an example time series in \cref{fig:pipeline_stacked}. In the middle panel, the knots are placed at equal curvature quantiles, resulting in a higher density in the complex region.
The bottom panel shows each basis function, flexible in scale and support, and the derived uniform-size BSAT tokens. 
For any number of underlying points, each token is a fixed-size tuple $(c_i,\mu_i)$, composed of two scalars: B-spline coefficient $c_i$ and center position of the basis function $\mu_i$. 
The center is constructed as  $\mu_i = \tfrac{1}{2}(\tau_i+ \tau_{i+p+1})$ and serves to encode positional information and implicitly encode knot density and local support width.
Consequently, BSAT can embed any variety of heterogeneous patch sizes into homogeneous tokens.
BSAT accommodates both sparsely and densely sampled time series, and can be applied to data with irregular sampling or missing observations. 
The spline-based approximation functions as a controllable low-pass noise filter with inductive smoothness and continuity bias. 
Moreover, BSAT allows precise control over the shape of the curve via the spline degree and the token budget.
BSAT is implemented  as a preprocessing step after mean-variance normalization. 
It fits a B-spline curve composed of $n$ basis functions to the value channels. 
This process reduces the number of input tokens from $L$ to $n$, decreasing attention layer complexity from $O(L^2)$ to $O(n^2)$.
For spline degree $p$ the value of $n$ can be chosen freely in the range $p+1 < n < L$, ensuring B-splines are both well-defined and uniquely determined \cite[Ch. 9]{de_boor_practical_2001}. 

Furthermore, BSAT incorporates the following components to ensure robustness and efficiency:

\paragraph{Modified Feature Function $f$ }
Operating on 1-D data allows us to modify Yeh et al.’s feature function \cite{yeh_fast_2020} to reduce numerical instability and lower rank deficiency risk: 
\[
f(\xi)=\bigl(|y^{(p)}(\xi)|+\varepsilon\bigr)^{1/p},\qquad
\varepsilon=10^{-6}.
\]
We (i) drop the square, unnecessary in 1-D; (ii) add $\varepsilon$ inside the root to ensure $f>0$ even on flat segments; (iii) use the $1/p$ exponent to avoid over-concentration of knots at sharp peaks, thus promoting more stable and balanced knot allocation.

\paragraph{Adaptive Clip-Factor} For low-degree, high-density splines  $g = 1$ may yield $\tilde w_i\!=\!w_i$ for many intervals, producing numerical rank deficiency.
Rank deficiency arises when the basis functions are nearly linearly dependent, rendering the design matrix singular or ill-conditioned. 
This prevents unique or stable least squares solutions.
Thus, we select $g$ heuristically for a given dataset via one-time grid search $g\in[0.10,1.25]$ with step 0.01 on the train fold: for each $g$ we fit $n$ degree 1 splines on $\mathcal{S}$ sliding windows with length $L$ and stride 100 and pick
\begin{equation}
g^* = \arg\min_{g}\; \max_{s \in \mathcal{S}} \mathrm{RMSE}(s,g).
\end{equation}
The number of search operations is $|\mathcal{G}|  \lceil\frac{ T_{train}}{ \text{stride}} \rceil$ with $\mathcal{G}$ number of grid points $g$, negligible relative to training.
\paragraph{Ridge Fallback} If the ratio of smallest to largest basis support is very large, the least square fit may become ill-conditioned, as basis vectors become collinear. 
Rather than abandoning the window, let $B \in \mathbb{R}^{ L\times n}$ denote the B-spline basis matrix evaluated at the window points and $y \in \mathbb{R}^{L}$ the observed values in the window. 
Let us define the Gram matrix $G = B^\top B$,  the $n \times n$ identity matrix $I$, $\mathrm{tr}(G)$ the trace of $G$,  the condition number $\kappa$ and $\lambda$  the Tikhonov (ridge) regularization parameter.
If $\kappa(G) > 10^8$, we fall back to solve the regularized system:
\[
(G + \lambda I)c = B^\top y, \qquad \lambda = 10^{-6}\,\frac{\mathrm{tr}(G)}{n}.
\]
This ensures a unique, numerically stable solution, but often at the cost of a higher error \cite[Ch. 6.1.4 - 6]{golub_matrix_2013}.

\paragraph{Coefficient Clipping} To prevent numerical instability in downstream operations, we clip coefficients:$|c_i| \leq C_{\max}$ for all $i$.

\paragraph{Cache} BSAT is deterministic for a dataset, given $(L,n,p, g)$. We calculate tokens once per run, then cache the tuples $(c_i,\mu_i)_{i=1}^n$, reducing per-epoch pre-processing to $O(1)$.
A pseudo-code implementation is available in \cref{app:bsat}.

\subsection{Positional Encoding for Non-Uniform Tokens}

\subsubsection{L-RoPE: Per-Layer Learnable Frequency Base}
RoPE uses a fixed $\theta_{\text{base}} = 10,000$ across all layers\cite{su_roformer_2023}, which imposes a rigid frequency spectrum. 
However, models use  different frequency dimensions to attend to position and semantic content respectively \cite{barbero_round_2025}. 
Additionally, contrary to natural language processing problems where a model must be able to operate on a diverse set of inputs, TSTs are trained on just one dataset, many of which exhibit significant recurring patterns.
Therefore, models may benefit from diverse RoPE frequency spectra, allowing them to intentionally attend to reoccurring dependencies at different temporal distances and adjust to the dataset's unique structure.
We introduce
\begin{equation}
\theta_{\text{base}}^{(l)} = \exp(\phi^{(l)}), \quad \phi^{(l)} \in \mathbb{R}
\end{equation}
where $\phi^{(l)}$ is a learnable log-scale parameter initialized to $\log(10,000)$ for layer $l$, guaranteeing a positive base and stable optimization, while maintaining the geometric progression of frequencies through the head. 

\subsubsection{Hybrid Additive Rotary Positional Encoding}
\begin{figure}[t]
  \centering
  \includegraphics[width=0.92\columnwidth]{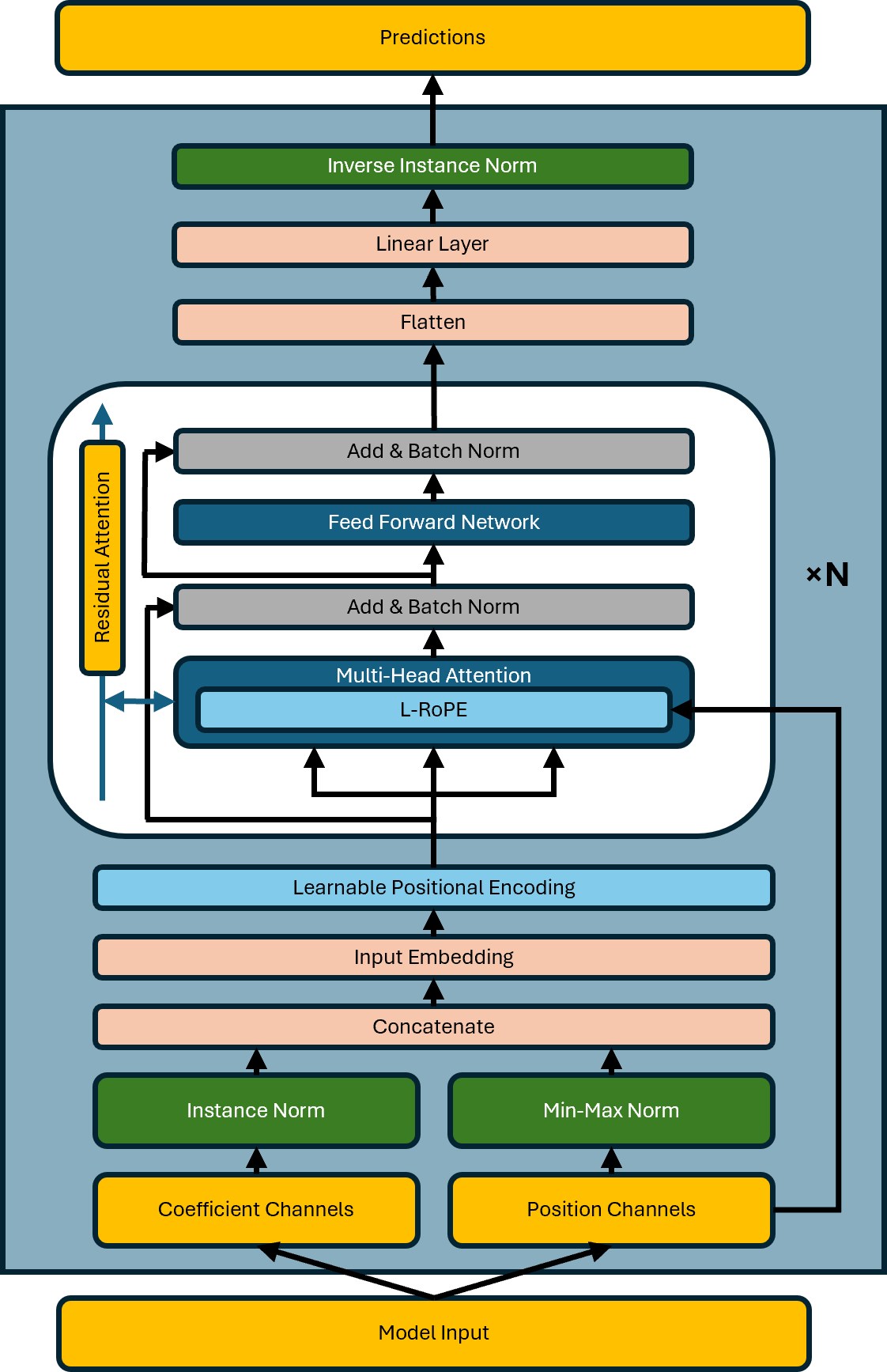}\\
 
  \caption{Model architecture: The network embeds coefficient and position channels, concatenates them, and applies both learnable positional encoding and RoPE. %RoPE is applied to the real-valued, non-normalized position channel: BSAT's basis centers $\mu$. 
  }
  \label{fig:arch}
\end{figure}

Traditional TSTs often rely on additive learned or sinusoid positional encodings. 
When RoPE is applied it is generally used mutually exclusively with positional encodings.
However, given the success of hybrid positional encodings for TSTs, we adopt a hybrid positional embedding scheme that embeds absolute position additively, and relative position rotatory. 
First an additive learned positional embedding (LPE) \cite{vaswani_attention_2017, DBLP:journals/corr/GehringAGYD17}
$\mathbf{e}^{\text{ord}}_{\,o(i)}\!\in\!\mathbb{R}^{d_{\text{model}}}$,
indexed by the left‑to‑right rank \(o(i)\), is added:
$
\mathbf{z}_i = E_{\text{in}}\mathbf{u}_i + \mathbf{e}^{\text{ord}}_{\,o(i)}.
$
Following that, relative distances between tokens are embedded with RoPE in the attention layer.  
This hybrid strategy address several concerns:  The large variance of token sizes may degrade RoPE performance by causing phases and, therefore, gradients to oscillate rapidly. The low-frequency additive LPE dampens this variance.  
Additionally, RoPE modifies only $Q$ and $K$, so absolute position would reach the residual path only via the attention weights; the additive term writes it directly into $V$, allowing downstream blocks to learn position-specific effects without first re-inferring them.  
Finally, the LPE guarantees each token's uniqueness, even if RoPE frequencies are repeated before the end of the window due to a small base.
In \cref{fig:arch}, the components of the hybrid positional encoding are shown in light blue.

\subsection{Model Architecture}
Model inputs consist of two distinct channel types: spline coefficients and center positions. Coefficient channels undergo reversible instance normalization \cite{kim_reversible_2021} to coefficients for distribution-invariant processing following \cite{Zeng_Chen_Zhang_Xu_2023, 10.1145/3534678.3539234,nie_time_2023}. Centers are min-max normalized based on $L$ to ensure consistent positional encoding despite adaptive token placement.
Channels are concatenated and embedded into a $d_{model}$-dimensional latent space via a trainable linear embedding layer.
Following PatchTST \cite{nie_time_2023} we employ a standard Transformer encoder architecture composed of $N$ layers. Each encoder layer features multi-head self-attention~\cite{vaswani_attention_2017} with residual connections~\cite{he_deep_2016}, followed by batch normalization~\cite{ioffe_batch_2015, zerveas_transformer-based_2021}, GELU activation~\cite{hendrycks_gaussian_2023}, and residual attention~\cite{he_realformer_2021}. Within each encoder layer, attention outputs are followed by a $d_{ff}$-dimensional feed-forward network defined as:
\begin{equation}
\mathrm{FFN}(x) = W_2(\mathrm{Dropout}(\mathrm{GELU}(W_1 x + b_1))) + b_2,
\end{equation}
where $W_1 \in \mathbb{R}^{d_{model} \times d_\mathrm{ff}}$, $W_2 \in \mathbb{R}^{d_\mathrm{ff} \times d_{model}}$ are weight matrices and $b_1 \in \mathbb{R}^{d_\mathrm{ff}}$, $b_2 \in \mathbb{R}^{d_{model}}$ are bias terms.
The encoder outputs are flattened along the token dimension and projected via a trainable linear head, producing the $H$ predictions.
These then undergo inverse scaling: first, inverse instance normalization, followed by de-normalization of global mean-variance normalization.  Normalization is fit on the training fold only.

\section{Experiments}
\paragraph{Datasets}
We benchmark on three public univariate series: ETTh1\footnote{\url{https://github.com/zhouhaoyi/ETDataset}} \cite{zhou_informer_2021}, Alabama PV 2006\footnote{\url{https://www.nrel.gov/grid/solar-power-data}} \cite{bloom2016eastern}, and ECL\footnote{\url{https://archive.ics.uci.edu/dataset/321/electricityloaddiagrams20112014}} \cite{trindade_electricityloaddiagrams20112014_2015}. 
Each is split chronologically 60\%/20\%/20\% (train/val/test).
Details, statistics, and preprocessing are found in ~\cref{app:datasets}.

\paragraph{Setup and Baseline} BSAT is compared to two common models that compress a 720 points time series into a token budget of  $T\!\in\!\{45,90,180\}$: A simple, \textit{Uniform Down Sampled Transformer (UDS)} and \textit{PatchTST} \cite{nie_time_2023} with $stride = \frac{720}{T}$ and $patch length = 2 \times stride$. 

\paragraph{Tuning and Training} 
We separately tune hyperparameters for each experiment over 200 runs via Bayesian optimization. 
This wide sweep is necessary as all models show significant sensitivity to hyperparameters, and BSAT lacks prior tuning guidance.
Training is conducted with a batch size of 128 and 100 epochs with early stopping and an asynchronous successive halving algorithm \cite{li_system_2020} and a cosine annealing learning rate scheduler with warm-up. 
All training uses a server with an A100 20GB MiG partition with FP32.
The complete search space, seeds, tuning and training configuration can be found in \cref{app:rep}.

\paragraph{Ablation Study}
To study the effect of the both the hybrid positional embedding and the learnable base, we conduct a ablation study (\cref{app:abl} across all tokenizers. We evaluate the hybrid embedding strategy against their respective pure RoPE variant: L-RoPE and F-RoPE (base 10.000).

\section{Results and Discussion}
\subsubsection{Comparative Benchmark}
\label{sec:main_results}
\begin{figure*}[t]
  \centering
  \includegraphics[width=0.9\textwidth]{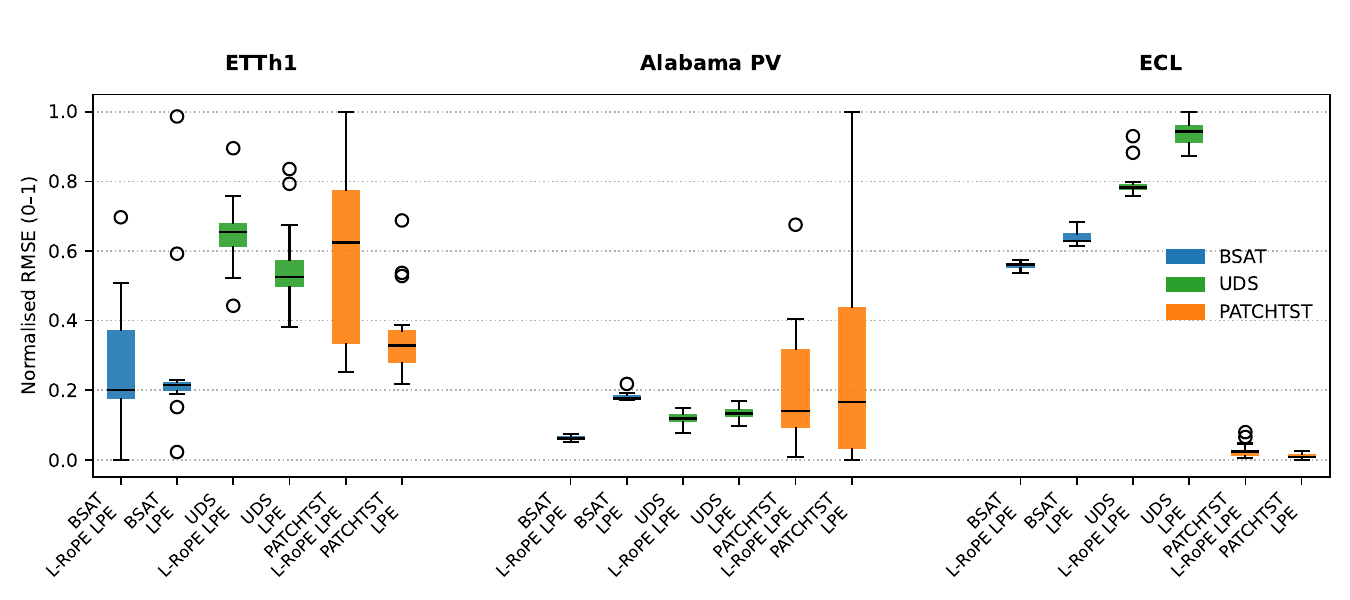}
  \caption{RMSE box-plot for token budget 45, showing the distribution of the top 15 runs for each configuration. RMSE is rescaled to $[0,1]$ inside each dataset~$\times$~budget group.}
  \label{fig:box_45}
\end{figure*}
Here we compare the performance of the best model configuration by average rank across all three datasets, all results based on the top 5 hyperparameter sets across 3 seeds. The full benchmark can be found in \cref{app:results}.

\textbf{Tokenization: }%BSAT -  is BSAT better than PatchTST/UDS? If so, under what conditions?
BSAT demonstrates strong results, dominating ETTh1 and delivering competitive results on the Alabama PV dataset.
In both datasets, it proves especially effective for low token budgets.
Contrary to UDS, it performed best on high budgets and is particularly successful on the Alabama PV data. 
Finally, PatchTST shines on ECL, where both UDS and BSAT struggle, due to the high variance of the data.
As outlined in Appendix \cref{app:datastats}, we use total variation and curvature (measured by the L2-Norm of the second derivative) to estimate the ease of accurately approximating a dataset with B-splines or down-sampled points: For both, ETTh1 measures the lowest and ECL the highest. 
This directly correlates with BSAT's relative performance. 
PatchTST, on the other hand, proves robust to this volatility, but fails to deliver top results on the other datasets. 
The difficult spline fitting on ECL results in BSAT repeatedly requires ridge regression fallbacks and produces large coefficients that must be clipped.
This never occurs on the other datasets.
When fitting the whole dataset with BSAT, using the heuristic derived $g$ ( ETTh1: 0.62, Alabama PV: 0.38, ECL: 1.25), only ECL produces coefficients larger than 10. 
Given mean coefficients ranging from -4 to 4, outlier coefficients in the hundreds (or thousands) can cause numeric instability in the model.
This observation suggests that the coefficient clipping heuristic and normalization schemes may be suboptimal for highly complex time series. 
The performance impact of this can be observed in the ECL BSAT L-RoPE LPE 180 box plot \cref{fig:box_180}, where we clearly see how the high token budget leads to higher volatility.
We further note that performance, for all models, is highly dataset dependent. 
None of the tokenizers evaluated here shows a consistently strong performance on all datasets. 
This supports the notion that time series tokenization constitutes a relevant field for further study as different data structures favor different tokenization strategies. 

\textbf{Embedding Methods: }% L-RoPE LPE - Is the specific combination of L-RoPE + LPE better than other embeddings? If so, under what conditions?
For both BSAT and UDS, the hybrid L-RoPE LPE embedding dominates traditional LPE, while for PatchTST, results are mixed.
For both PatchTST and UDS, LPE outperforms L-RoPE LPE on ETTh1, and for all models, L-RoPE LPE beats LPE on the Alabama PV dataset. 
On ECL results are mixed for PatchTST, in favor of L-RoPE LPE for UDS, and in favor of LPE for BSAT.   
Considering the relative change from LPE to L-RoPE, we observe a mean RMSE change of -3.13\% for BSAT, -2.23\% for PatchTST, and -0.1\% for UDS.
For all models, the strongest effects are observed on Alabama PV, where on the 180 token configuration BSAT L-RoPE LPE shows a -10.1\% and PatchTST L-RoPE LPE shows a -15.45\% improvement over their LPE counterparts. 

\textbf{Token Budgets \& Compute Efficiency: }As illustrated by \cref{fig:box_45}, \cref{fig:box_90} and \cref{fig:box_180}, BSAT performs strongest on the lowest token budget. On ETTh1, it achieves lower minimum, mean, and median RMSE than UDS or PatchTST, even at higher token budgets.  
Similarly strong performance on the $T=45$ token budget is achieved on Alabama PV.  
This reduced input representation results in substantial improvements to computational efficiency: reducing the token budget from 180 to 45 decreases peak GPU memory usage by a factor of 8.
On ETTh1, it drops from 1200 MiB to just 150 MiB.
Achieving this efficiency without deteriorating performance makes BSAT particularly well-suited for resource-constrained environments and long-sequence forecasting.

\subsection{Ablation Results}
\label{abl}
In this section we discuss the ablation results as reported in \cref{app:abl}.

\subsubsection{Hybrid Additive Rotary Positional Encoding RoPE LPE}
We observe that for BSAT supplementing RoPE variants with an absolute Learned Positional Encoding (LPE) consistently improves performance, beyond gains from L-RoPE alone. 
This finding is at odds with prior work \cite{zivanovic_rotary_2025} where absolute positions could be recovered from RoPE alone. 
We attribute this difference to the non-uniform, overlapping nature of BSAT's tokens, which may make it harder for the model to infer absolute order from relative rotations alone, thus making an explicit absolute encoding like LPE beneficial.
Notably we show that, while results are mixed, the hybrid embeddings also improve the performance of uniform tokenizer models, especially on UDS, where it was able to outperform the LPE embedding.
We often observed a substantial improvement in $\sigma$ across all models, when comparing hybrid models to their pure RoPE counterparts and LPE.

\subsubsection{L-RoPE as a Multi-Resolution Mechanism}
\label{RoPE}
\begin{figure}[t]
  \centering
  \includegraphics[width=0.92\columnwidth]{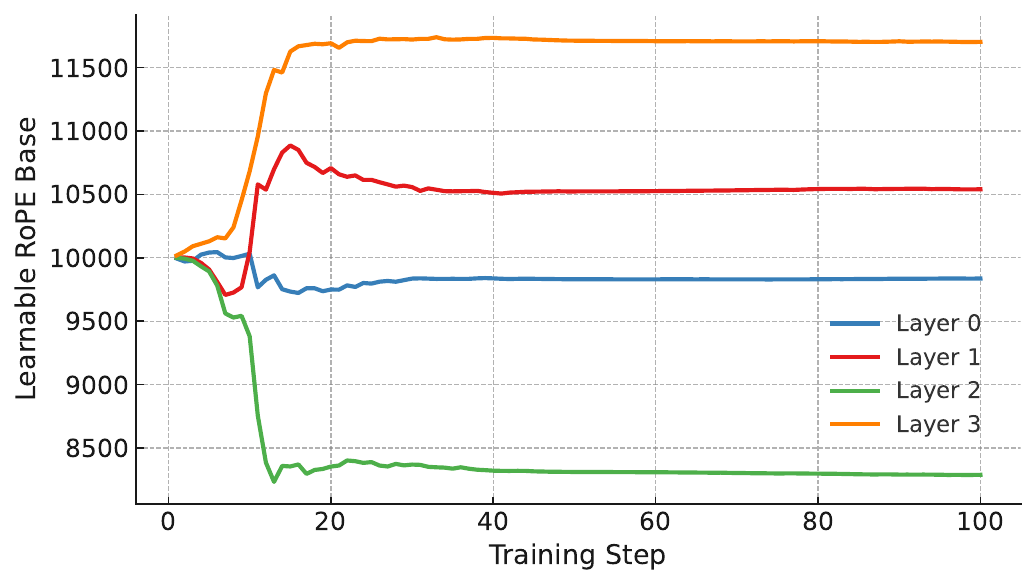}
  \caption{RoPE base layer convergence for one BSAT L-RoPE LPE run (16 dimensions, 4 heads) on Alabama PV. Layers specialize in different frequencies in an alternating pattern.}
  \label{rope_example}
\end{figure}

During training, we observed that different layers consistently learn to specialize their RoPE bases to distinct values, as illustrated in \cref{rope_example}. 
Rather than converging to a single optimal base for a given dataset, the layers tend to move in opposing directions, develop a diverse set of base values.
There is no configuration of base values that appears ideal, but several pattern configurations with similar performance.
Rather than due to specific base values, these results suggest performance gains from L-RoPE originate from this learned diversity in base values. 
In the appendix heatmap (\cref{fig:rope}), this diversity is quantified and visualized how the benefit of this learned spread is highly dataset-dependent. 
We theorize that by equipping layers with different frequency sensitivities, the model can influence its attention pattern, similarly to how multi-head attention attends to different dependencies.
This allows layers to specialize in attending to either short-range or long-range dependencies. 
The performance improvements are strongest on the Alabama PV dataset, which contains strong periodic patterns.
Notably here we see the biggest performance gains in the highest token budget, while for ECL gains are exclusively found for lower token counts.
ETTh1 on the other hand has the weakest periodic patterns and only shows gains for BSAT.
This implies that L-RoPEs efficiency is not just heavily dependent on the underlying data patterns, but also on the tokenization method and compression.
Additionally we find that bases change more on hybrid than on pure embeddings.
We theorize that L-RoPE serves a fundamentally different role here than RoPE traditionally would, and this role is enhanced through its application together with learned positional embeddings (LPE). 
Since LPE already encodes absolute positions, L-RoPE is freed from its traditional positional encoding role and instead functions as an attention pattern modulator.
On datasets with strong reoccurring patterns, the model can then deliberately attend to those tokens.
Bases are often adjusted no more than 20\% from its initialization of 10.000, but given the volatile attention allocation produced by RoPE, even small base changes enable to model to shift its attention significantly.
%The exact mechanism by which the learned RoPE base interacts with temporal patterns and influences performance requires further study.

Overall, we consider both L-RoPE and the L-RoPE LPE hybrid positional encoding to be a meaningful contribution to the growing RoPE literature. Implementation and parameter cost are trivial, and the ability to modulate attention patterns to temporal structure offers a simple yet effective means to encode latent temporal priors, enabling the model to uncover and leverage non-obvious temporal dependencies without added complexity.

\subsection{Limitations and Diagnostics}
While BSAT demonstrates strong performance and efficiency, our analysis reveals one notable limitation: volatile datasets can cause numerical instability.
This challenge arises from the B-spline fitting process, particularly at high token densities and on datasets with high total variation like ECL (\cref{app:datastats}). 
On such data, the fitting algorithm struggles to maintain numerical stability, forcing it to frequently employ ridge regression fallbacks, leading to worse approximations and producing large B-spline coefficients that must be clipped to prevent model divergence, losing information. 
This information loss is particularly acute for segments with rapid oscillations or significant outliers, leading to over-smoothing and debilitatingly large coefficients. 
This becomes evident on ECL, the only dataset where BSAT shows substantial, consistent performance degradation with higher token budgets, as shown in \cref{fig:box_180}.
This indicates that our heuristic for clip factor $g$, coefficient clipping value, and normalization schemes are insufficient.
In line with this diagnostic analysis, higher spline degrees, which naturally have a larger support, consistently outperform lower ones, suggesting that while model flexibility is crucial, it must be paired with robust regularization to be effective.

\subsection{Future Work}
To address the numerical stability of the BSAT, we propose two improvements: (i) modifying the knot placement algorithm to enforce a maximum base support ratio, guaranteeing a well-conditioned least-squares fit without requiring a ridge fallback, and (ii) applying a inverse hyperbolic sine transformation to coefficients to manage their dynamic range without clipping.
Inspired by \cite{saillot_b-spline_2024}, a more ambitious direction is end-to-end differentiability for the entire tokenization process via a learnable B-spline fitting model.
Further, the fitting strategy itself can be enhanced. Drawing inspiration from recency-weighted models \cite{johnsen_recency-weighted_2024}, the knot placement algorithm could be modified to enforce a higher density of knots on more recent data, adding a recency bias.
Finally the RoPE base divergence and its interaction with LPE in the hybrid embedding show an notable patter that is not explained well by existing RoPE literature, offering a possible topic for further study.

\section{Conclusion}
We address two major challenges of contemporary TSTs: Computational inefficiency and rigid patch segmentation. We introduced the B-Spline Adaptive Tokenizer, a parameter-free method that aligns tokenization with the semantic structure of a time series by placing tokens densely in high-curvature regions. To complement this, we developed a hybrid positional encoding strategy with a layer-wise learnable frequency base for RoPE, designed to accurately represent the position of these non-uniform tokens.

Our primary finding is that adaptive, data-driven tokenization via BSAT enables superior forecasting performance, especially at high compression ratios where it becomes more computationally efficient. We demonstrate that the attention layers successfully learn a spectrum of frequency bases, effectively creating a multi-resolution attention mechanism.
Future work might focus on improving the robustness of the spline fitting algorithm for highly volatile time series.

\clearpage
\appendix
\section{Appendix}

\subsection{Dataset Details}
\label{app:datasets}
See \cref{tab:datasets} for full specifications of each dataset.
\begin{table*}[t]
\centering
\small
\setlength{\tabcolsep}{6pt}
\begin{tabular}{p{1.5cm}ccccp{3.6cm}cp{2cm}}
\toprule
Dataset & Cadence & Raw Len & Used Len & Split (Train/Val/Test) & Target Tag & Unit & Preprocessing \\
\midrule
ETTh1 & 1\,h & 17{,}420 & 17{,}420 & 10{,}452 / 3{,}484 / 3{,}484 & OT & $^\circ$C & None. \\
Alabama PV 2006 & 5\,min & 105{,}120 & 35{,}040 & 21{,}024 / 7{,}008 / 7{,}008 & Actual\_35.05\_-87.65\_2006 \_DPV\_38MW\_5\_Min& MW & Aggregated to 15\,min (mean). \\
ECL & 15\,min & 140{,}256 & 70{,}176 & 42{,}106 / 14{,}035 / 14{,}035 & MT\_320 & kW & First 2 years kept. \\
\bottomrule
\end{tabular}

\caption{Full dataset specifications.}
\label{tab:datasets}
\end{table*}

\subsubsection{Dataset Analysis}
\label{app:datastats}
Given the divergence in model performances across datasets, we evaluated them with total variation, L2 norm of the second derivative, and Permutation Entropy \cite{tibshirani_adaptive_2014, eilers_flexible_1996}.
Total variation $\sum_{i} |x_{i+1} - x_i|$ measures cumulative point-to-point change, while the L2 norm of second differences $\left\|\nabla^2 x\right\|_2$ quantifies local curvature; both are applied as B-spline fitting complexity metrics. Permutation entropy $H_p$ captures ordinal pattern diversity, indicating intrinsic forecasting difficulty due to temporal structure randomness.
The results can be seen in \cref{tab:ts_stats}.

Our results (\cref{app:results}) show that BSAT performance correlates inversely with both total variation and L2 norm, indicating sensitivity to dataset complexity. UDS performance exhibits its strongest correlation with permutation entropy. PatchTST demonstrates robustness to geometric complexity metrics but shows inverse correlation with permutation entropy, notably under-performing on the Alabama PV dataset.
\begin{table}
\centering
\small
\begin{tabular}{l p{1.2cm} c p{2cm}}
\toprule
Dataset &  Total Variation& L2 Norm& Permutation Entropy\\
\midrule
ETTh1 & 10699 & 172 &0.92\\
Alabama PV& 21696& 223& 0.51\\
ECL& 51268& 713& 0.83\\
%ETTm2 & 15860 & 113 & 0.67 \\
\bottomrule
\end{tabular}
\caption{Comparison of total variation, L2 norm of second derivative, and permutation entropy for all datasets.}
\label{tab:ts_stats}
\end{table}

\subsection{Reproducibility}
\label{app:rep}
We ensure reproducibility using fixed random seeds {2025, 2026, 2027} and PyTorch 2.5 with deterministic algorithms across all experiments. 
For each seed, the best 5 runs by validation RMSE are selected, and metrics are reported as the average across these 15 runs. 
%The model is implemented within the %CommonPower\cite{eichelbeck_commonpower_2025} code base.
%maybe remove to ensure anonymity

\subsubsection{Hyperparameter Tuning Configuration}
\label{app:space}
\paragraph{Search Strategy:} To effectively optimize the complex search space, we conduct hyperparameter optimization using Ray Tune 2.41 with the Optuna 4.2.1 Tree-structured Parzen Estimator sampler, configured with 40 startup trials, 500 Expected Improvement (EI) candidates, with the constant-liar, multivariate, and group settings enabled. This approach offers greater sample efficiency and flexibility compared to random or grid search, particularly in high-dimensional or conditional parameter spaces.Each configuration explores 200 trials using an asynchronous successive halving scheduler with reduction factor 3 and grace period of 20 epochs. Early stopping combines a maximum of 100 epochs and patience-based stopping, monitoring improvements in validation RMSE for 10 epochs with a grace period of 10. We employ 5 concurrent trials.
\paragraph{Search Space:} 
The search space includes: transformer layers (2-4), model dimension $d_\mathrm{model} \in \{16, 32, 64, 128\}$, attention heads $h \in \{4, 8\}$, feed-forward factor $d_{\text{ff}\_\text{factor}} \in \{2, 4, 6, 8\}$ with $d_\mathrm{ff} = d_\mathrm{model} \times d_{\text{ff}\_\text{factor}}$.
General dropout rate p is sampled from [0, 0.4], fully-connected dropout from [0, 0.2], and attention dropout from [0, 0.4], all are shared across layers.
Learning rate is chosen log-uniform [1e-6, 1e-3], and weight decay log-uniform [1e-6, 1e-2]. BSAT additionally tunes the spline degree in \{1, ..., 6\}.

\subsubsection{Training Configuration}
Models are trained using AdamW optimization with Xavier uniform initialization and gradient clipping at L2-norm of 1.0. The learning rate follows a cosine annealing schedule that begins at 5\% of the sampled learning rate, warms up linearly to the full rate over 10 epochs, then decays to 1\% of the peak rate over 40 epochs. We optimize using MSE loss for stable gradients, while all model selection and early stopping decisions use RMSE for interpretability. 

\subsubsection{Pseudo Code}
\label{app:bsat}
We provide pseudo code for the generation of the knot vector following \cite{yeh_fast_2020} in \cref{alg:akp} and BSAT in \cref{alg:bsat}.
% ------------- Alg. 1 -------------
\begin{algorithm}[t]
\small
\caption{Adaptive Knot Placement\label{alg:akp}}
\begin{algorithmic}[1]
\Require $\xi_{0:L-1},\;y_{0:L-1},\;p,\;n,\;g$
\Ensure  $\boldsymbol{\tau}$
\State $dy\gets y$
\For{$i=1$ \textbf{to} $p$}                           \Comment{$p$-th derivative}
    \State $dy\gets \operatorname{gradient}(dy,\xi)$
\EndFor
\State $\eta\gets10^{-6}\,\text{mean}(|dy|)$
\State $\varphi_\ell\gets(|dy_\ell|+\eta)^{1/\max(p,1)}$
\State $\omega_\ell\gets\frac12(\varphi_{\ell+1}+\varphi_\ell)(\xi_{\ell+1}-\xi_\ell)$
\State $k_{\mathrm{int}}\gets k-2(p+1)$ \Comment{$k=n+p+1$}
\State $\Delta F\gets\sum\omega_\ell/k_{\mathrm{int}}$
\State $\omega_\ell\gets\min(\omega_\ell,\,g\Delta F)$     \Comment{clipping}
\State $F\gets[0,\;\text{cumsum}(\omega)];\;F\gets F/F_{-1}$
\State $\xi^{\text{mid}}_\ell\gets
       \begin{cases}\xi_0,&\ell=0\\
       \tfrac12(\xi_{\ell-1}+\xi_\ell),&\ell>0\end{cases}$ \Comment{mid-points}
\State $q\gets\text{linspace}(0,1,k_{\mathrm{int}}+2)_{1:{-}1}$ \Comment{equal-mass quantiles}
\State $\text{interior}\gets\text{interp}(q;\,F\leftrightarrow\xi^{\text{mid}})$
\State $\boldsymbol{\tau}\gets
       \underbrace{\xi_0,\dots,\xi_0}_{p+1}\,\Vert\;\text{interior}\;\Vert\,
       \underbrace{\xi_{L-1},\dots,\xi_{L-1}}_{p+1}$
\State \Return $\boldsymbol{\tau}$
\end{algorithmic}
\end{algorithm}

% ------------- Alg. 2 -------------
\begin{algorithm}[t]
\small
\caption{BSAT \label{alg:bsat}}
\begin{algorithmic}[1]
\Require $\mathbf{X}\!\in\!\mathbb{R}^{B\times L\times C},\;p,\;n,\;g,\;c_{\max}$
\Ensure  $\hat{\mathbf{C}}\!\in\!\mathbb{R}^{B\times n\times C},\;\boldsymbol{\mu}$
\State $\xi_{0:L-1}\gets\text{linspace}(0,1,L)$            \Comment{shared grid}
\For{$b=1$ \textbf{to} $B$}
  \For{$c=1$ \textbf{to} $C$}
    \State $\mathbf{y}\gets\mathbf{X}[b,:,c]$
    \State $k\gets n+p+1$
    \State $\boldsymbol{\tau}\gets$ Alg.~\cref{alg:akp}$(\xi,\mathbf{y},p,n,g)$
    \State \textbf{try}\;
      $\hat{\mathbf{c}}\gets
      \operatorname*{arg\,min}_{\mathbf{c}}\lVert
      \mathbf{y}-\mathbf{N}(\boldsymbol{\tau})\mathbf{c}\rVert_2^2$   \Comment{LSQ fit}
    \State \textbf{catch}\;
      $\mathbf{N}_{\ell,i}\gets N_{i,p}(\xi_\ell)$,
      $\mathbf{G}\gets\mathbf{N}^\top\mathbf{N}+10^{-6}\mathbf{I}$,
      $\hat{\mathbf{c}}\gets\mathbf{G}^{-1}\mathbf{N}^\top\mathbf{y}$ \Comment{ridge fallback}
    \State $\hat{\mathbf{C}}[b,:,c]\gets\hat{\mathbf{c}}$
  \EndFor
  \State $\mu_i\gets\tfrac12(\tau_i+\tau_{i+p+1})\times(L-1)$ \Comment{centers}
\EndFor
\State $\hat{\mathbf{C}}\gets
       \text{clip}(\hat{\mathbf{C}},-c_{\max},c_{\max})$
\State \Return $(\hat{\mathbf{C}},\;\boldsymbol{\mu})$
\end{algorithmic}
\end{algorithm}

\subsection{Ablation}
\label{app:abl}
On both the benchmark \cref{app:results} and the ablation results \cref{tab:etth1_ablation}, L-RoPE LPE showed very competitive performance. 
Across 9 dataset-token budget combinations, BSAT L-RoPE LPE outperforms BSAT LPE 7 times, UDS L-RoPE LPE outperforms UDS LPE 6 times, and PatchTST L-RoPE LPE outperforms PatchTST LPE 4 times and ties once.

We observe that BSAT strongly benefits from the hybrid L-RoPE LPE across all token budgets, consistently outperforming both pure relative (L-RoPE, F-RoPE) and pure absolute (LPE) embeddings. 
While UDS and PatchTST also improved performance over their respective LPE models, they did so on other datasets.
BSAT saw improvements for all token budgets on ETTh1 \cref{tab:appendix_results_ci_etth1}, while UDS and PatchTST saw none. 
On ECL \cref{tab:appendix_results_ci_ecl} on the other hand, both UDS and PatchTST saw very strong results with L-RoPE LPE, while it performed the worst out of all datasets for BSAT. 
Notably on the Alabama PV \cref{tab:appendix_results_ci_al_solar06}, characterized by strong periodic patterns, hybrid embeddings universally enhanced performance across tokenizers. 
A possible explanation is that because the dataset spans only a single year and daylight length changes sharply with season, additive LPE struggles with identical token indices that now encode different seasonal patterns. In contrast, L-RoPE can shift its base frequencies, reshaping the attention kernel to attend a broader range of temporal dependencies and thus remains more robust when confronted with out-of-distribution patterns. 
This suggests that L-RoPE LPE's advantages depend on temporal data structures, and are not strictly linked to tokenization strategy alone.

To quantify performance stability, we additionally present a Coefficient of Variation ($CV = \frac{\sigma}{\mu}$) table \cref{tab:appendix_cv} for the benchmark models. 
A low CV implies stable, robust model performance and good generalization, essential for avoiding overfitting or brittle solutions and managing concept drift.

 BSAT L-RoPE LPE exhibits the lowest CV overall, with BSAT LPE and BSAT F-RoPE LPE second and third. 
Additionally, we observe a spike in CV for BSAT LPE, F-RoPE, and L-RoPE at token budget 90 on ETTh1. Notably, this spike is absent in the BSAT hybrid embeddings, suggesting that by hybridizing the embeddings, we achieve robustness to variations in data structure that would degrade the performance of the pure embedding methods. 
However, this stability improvement cannot be exclusively attributed to hybrid embeddings; rather, we show BSAT to be a remarkably stable tokenizer that is further enhanced with L-RoPE LPE.
When compared to UDS and PatchTST, where for each configuration at least one CV is 10\% or higher, the largest CV for BSAT LPE is 6.7\% and for BSAT L-RoPE LPE is 4\%.
We argue that this is evidence that BSAT successfully supports effective attention allocation by placing denser tokens in higher complexity regions, allowing the model to more reliably extrapolate to unseen patterns.
While pure RoPE strategies struggle without the explicit encoding of absolute order -- a critical feature for TSTs --  the combination of L-RoPE with LPE allows the layers to specialize in different temporal distances, which improves both performance and stability. This is because, due to the non-uniform tokenization, recurring patterns can no longer be inferred from LPE alone.

\begin{table*}[htbp]
\centering
\begin{tabular}{l|ccc}
\toprule
\textbf{Model} & \textbf{45} & \textbf{90} & \textbf{180} \\ \midrule
BSAT LPE & \underline{2.944±0.103} & 3.007±0.192 & 2.986±0.059 \\
BSAT F-RoPE & 2.965±0.098 & 3.040±0.292 & 3.040±0.217 \\
BSAT L-RoPE & 2.962±0.103 & 3.128±0.341 & \underline{2.978±0.049} \\
BSAT F-RoPE LPE & 2.964±0.105 & \underline{2.973±0.103} & 2.999±0.064 \\
BSAT L-RoPE LPE & \textbf{2.942±0.080} & \textbf{2.947±0.083} & \textbf{2.977±0.054} \\
\midrule
UDS LPE & 3.075±0.055 & 3.055±0.065 & 2.989±0.061 \\
UDS F-RoPE & 3.077±0.045 & 3.199±0.152 & 3.079±0.227 \\
UDS L-RoPE & 3.155±0.068 & 3.180±0.189 & 3.106±0.232 \\
UDS F-RoPE LPE & 3.146±0.103 & 3.197±0.160 & 3.121±0.133 \\
UDS L-RoPE LPE & 3.113±0.048 & 3.135±0.041 & 3.084±0.137 \\
\midrule
PATCHTST LPE & 2.984±0.059 & 3.001±0.025 & 2.978±0.053 \\
PATCHTST F-RoPE & 3.057±0.118 & 3.039±0.100 & 3.079±0.208 \\
PATCHTST L-RoPE & 3.078±0.142 & 3.035±0.073 & 2.987±0.079 \\
PATCHTST F-RoPE LPE & 3.012±0.075 & 3.035±0.076 & 2.986±0.023 \\
PATCHTST L-RoPE LPE & 3.083±0.117 & 3.040±0.056 & 3.000±0.031 \\
\bottomrule
\end{tabular}
\caption{Ablation results for RMSE on ETTh1 dataset across different token budgets. Shows Mean ± Standard deviation for all experiments. Best results in bold, second best underlined.}
\label{tab:etth1_ablation}
\end{table*}

\begin{table*}[htbp]
\centering
\small
\begin{tabular}{lcccccccccc}
\toprule
& \multicolumn{3}{c}{ETTh1} & \multicolumn{3}{c}{AL Solar06} & \multicolumn{3}{c}{ECL} & \multirow{2}{*}{\textbf{Average CV}} \\ 
\cmidrule(lr){2-4}\cmidrule(lr){5-7}\cmidrule(lr){8-10}
\textbf{Model} & 45 & 90 & 180 & 45 & 90 & 180 & 45 & 90 & 180 &  \\ 
\midrule
BSAT LPE & 3.5\% & 6.4\% & 2.0\% & 0.9\% & 2.0\% & 6.7\% & 0.5\% & 0.1\% & 0.1\% & \underline{2.5\%} \\
BSAT F-RoPE & 3.3\% & 9.6\% & 7.1\% & 1.8\% & 3.4\% & 3.5\% & 0.3\% & 1.2\% & 0.9\% & 3.5\% \\
BSAT L-RoPE & 3.5\% & 10.9\% & 1.7\% & 0.8\% & 2.5\% & 5.1\% & 0.2\% & 1.9\% & 1.9\% & 3.2\% \\
BSAT F-RoPE LPE & 3.5\% & 3.5\% & 2.1\% & 6.8\% & 2.3\% & 4.0\% & 0.3\% & 1.1\% & 2.6\% & 2.9\% \\
BSAT L-RoPE LPE & 2.7\% & 2.8\% & 1.8\% & 0.6\% & 2.8\% & 3.4\% & 0.3\% & 1.2\% & 4.0\% & \textbf{2.2\%} \\
\midrule
UDS LPE & 1.8\% & 2.1\% & 2.0\% & 1.7\% & 10.4\% & 5.4\% & 0.9\% & 1.6\% & 4.9\% & 3.4\% \\
UDS L-RoPE LPE & 1.5\% & 1.3\% & 4.4\% & 1.5\% & 7.7\% & 10.1\% & 1.3\% & 1.8\% & 2.1\% & 3.5\% \\
\midrule
PATCHTST LPE & 2.0\% & 0.8\% & 1.8\% & 19.1\% & 16.9\% & 18.0\% & 0.2\% & 0.2\% & 0.1\% & 6.6\% \\
PATCHTST L-RoPE LPE & 3.8\% & 1.8\% & 1.0\% & 13.3\% & 17.4\% & 17.1\% & 0.6\% & 0.2\% & 0.1\% & 6.2\% \\
\bottomrule
\end{tabular}
\caption{Coefficient of Variation (Std/Mean) of RMSE for all datasets and token budgets. Lower values indicate more stable performance. The final column shows the average CV across all tasks, best result bold, second best underlined.}
\label{tab:appendix_cv}
\end{table*}

\subsubsection{Learned RoPE Base Heatmap}
\label{app:rope}
\begin{figure*}
  \centering
  \includegraphics[width=\textwidth]{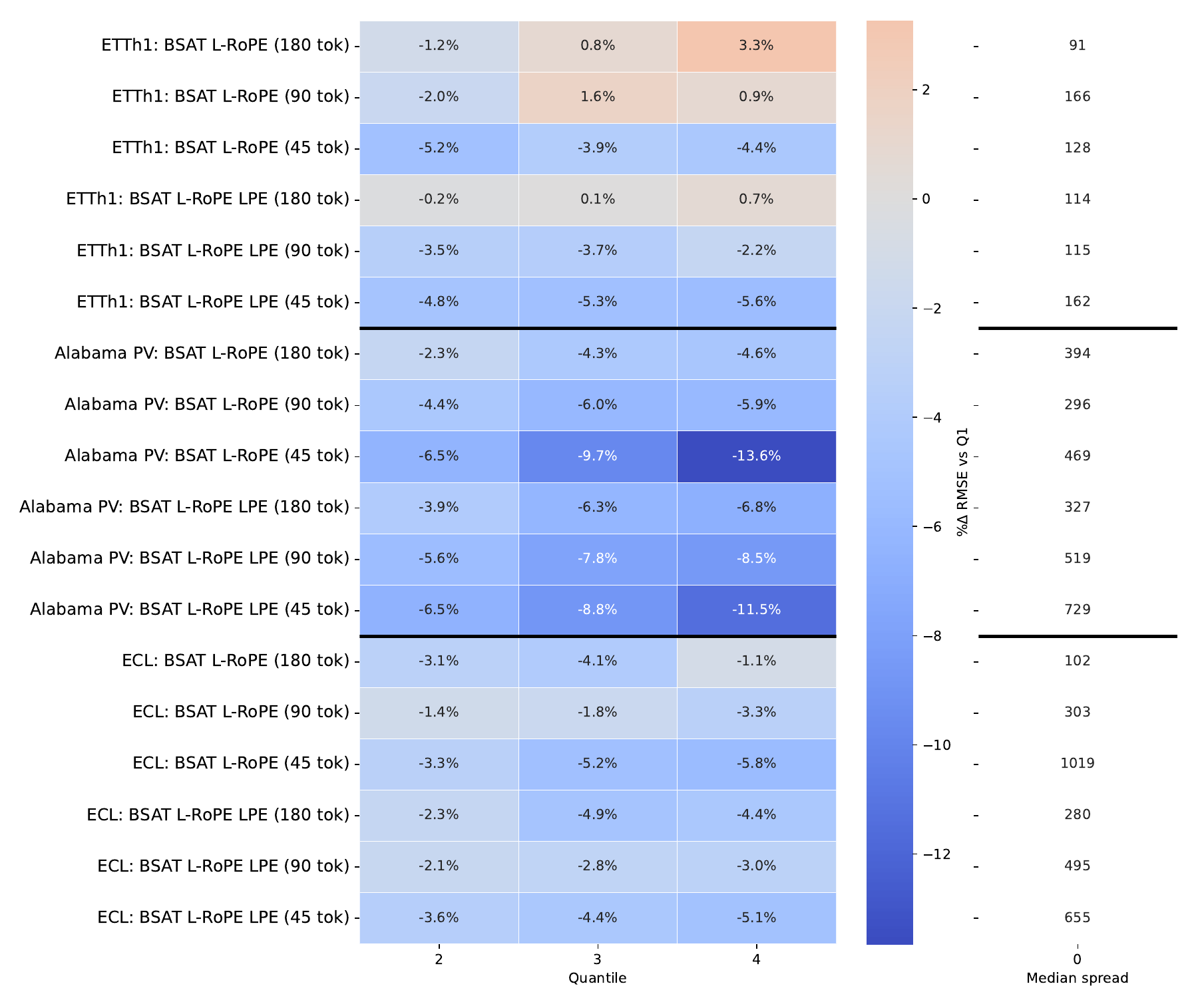}
  \caption{Effect of learned RoPE base diversity. Cells show test RMSE change relative to the tightest-spread quartile (Q1) across quartiles Q2–Q4. Each cell shows the percent change in mean RMSE, negative values (blue) indicate performance improvements. The rightmost column shows the overall median spread per row. Wider spreads markedly help on Alabama PV and ECL and offer modest gains on ETTh1; the benefits diminishes as the token budget grows. Calculated on all runs.}
  \label{fig:rope}
\end{figure*}

The heatmap in \cref{fig:rope} shows RMSE results for all 600 runs per experiment, for all token budgets, quartiled based on the L-RoPE and L-RoPE LPE base spread (difference between minimum and maximum base values) at the time of run termination.
It further shows the median RoPE spread per experiment.
This reveals the extent to which the model allows RoPE base values to diverge per configuration and how this divergence impacts model performance.
We observe that RoPE-spread effects and median spread vary strongly by dataset and token budget. 

For Alabama PV, in both L-RoPE and L-RoPE LPE, increased RoPE-spread leads to significantly lower RMSE, with median spreads rising substantially as token budgets decrease. 
ETTh1 shows similar, but weaker, more ambiguous trends, here for $T = 180$ higher quartiles show a slight increase in RMSE compared to the first quartile.
Finally for ECL we see consistent, moderate improvements for both L-RoPE and L-RoPE LPE.
As is for all datasets, these performance gains grow as the token budgets decrease.
The only notable outliers are the fourth quartile for  $T=180$, there however BSAT experienced difficulties as discussed in \cref{sec:main_results}, and results from associated experiments may not be representative.

Notably, the benefit of spread is most pronounced at the lowest token budget.
We theorize that as the model has access to fewer tokens, each of these tokens represents larger, impactful semantic structures, and the model must thus carefully control which temporal distances to attend to, to maximize the limited information available. 
Each base frequency introduces specific positive or negative attention biases toward tokens.
Assuming that attending to some tokens improves performance while attending to others hinders it, these bases become more effective and easier to optimize, as the token count decreases.

This also aligns with the comparison of L-RoPE and L-RoPE LPE performance: while L-RoPE often sees higher gains at the lowest token budget, at medium and high token budgets L-RoPE LPE appears to perform better. 
For L-RoPE FFNs, the only way to attain absolute positional information is to reconstruct it based on the relative distances embedded in the self-attention.
This is not trivial and becomes harder as more and denser tokens are created. 
For L-RoPE LPE, on the other hand, tokens already contain positional information. However, these embeddings still influence attention allocation statically across all layers, somewhat diluting L-RoPE's ability to modulate each layer's attention patterns. 
However, on higher token budgets, with important semantic patterns composed of several dense tokens, this positional bias becomes beneficial.
While for L-RoPE, given its limited capacity, absolute positions become harder to reconstruct and it struggles to find advantageous frequencies as a reversion to the mean effect takes hold, L-RoPE LPE, while also subject to this effect, it can use the learned positional embedding to maintain absolute order and maintain more consistent attention patterns, leading to stronger performance.

\subsection{Full Benchmark}
\label{app:results}
 The complete results are listed in: \cref{tab:appendix_results_ci_etth1}, \cref{tab:appendix_results_ci_al_solar06} and \cref{tab:appendix_results_ci_ecl}. 
For all metrics we report means with 95\,\% bias-corrected and accelerated bootstrap (BCa) confidence intervals: 
end-points $\hat\theta_{(\alpha_{1,2})}$ use $\alpha_{1,2}=\Phi\!\bigl(z_0+\frac{z_0\pm z_{0.975}}{1-a\,(z_0\pm z_{0.975})}\bigr)$, 
where $z_0=\Phi^{-1}\!\bigl[\Pr(\hat\theta^*<\hat\theta)\bigr]$ and $a$ is the jack-knife acceleration, computed with $B=10{,}000$ resamples. 
Non-overlapping BCa bands imply the models’ mean forecast errors differ beyond resampling noise; broader bands indicate less stable predictions and higher hyperparameter sensitivity.

\begin{table*}[htbp]
\centering
\small
\begin{tabular}{llccc}
\toprule
\textbf{Model} & \textbf{Metric} & \textbf{45} & \textbf{90} & \textbf{180} \\ \midrule
\multirow{4}{*}{BSAT LPE} & RMSE & \underline{2.944 (2.910/3.031)} & 3.007 (2.940/3.145) & 2.986 (2.958/3.016) \\
 & MAE & \underline{2.318 (2.287/2.407)} & 2.368 (2.310/2.483) & 2.353 (2.331/2.378) \\
 & MSE & \underline{8.677 (8.465/9.221)} & 9.075 (8.642/9.993) & 8.917 (8.754/9.099) \\
 & SMAPE & \underline{29.531 (29.149/30.599)} & 30.562 (29.592/32.402) & 30.024 (29.751/30.329) \\
\cmidrule(lr){2-5}
\multirow{4}{*}{BSAT L-RoPE LPE} & RMSE & \textbf{2.942 (2.910/2.990)} & 2.947 (2.914/2.997) & 2.977 (2.957/3.016) \\
 & MAE & \textbf{2.318 (2.292/2.364)} & 2.333 (2.303/2.384) & 2.347 (2.333/2.369) \\
 & MSE & \textbf{8.661 (8.466/8.958)} & 8.691 (8.497/9.006) & 8.865 (8.748/9.109) \\
 & SMAPE & \textbf{29.525 (29.135/30.084)} & 29.593 (29.241/30.123) & 29.859 (29.655/30.256) \\
\midrule
\multirow{4}{*}{UDS LPE} & RMSE & 3.075 (3.052/3.108) & 3.055 (3.023/3.085) & 2.989 (2.961/3.020) \\
 & MAE & 2.455 (2.440/2.475) & 2.420 (2.398/2.442) & 2.379 (2.357/2.401) \\
 & MSE & 9.456 (9.319/9.662) & 9.338 (9.142/9.529) & 8.938 (8.768/9.125) \\
 & SMAPE & 31.209 (30.977/31.475) & 30.841 (30.474/31.206) & 30.353 (30.024/30.695) \\
\cmidrule(lr){2-5}
\multirow{4}{*}{UDS L-RoPE LPE} & RMSE & 3.113 (3.091/3.137) & 3.135 (3.117/3.158) & 3.084 (3.026/3.160) \\
 & MAE & 2.490 (2.469/2.509) & 2.485 (2.471/2.503) & 2.443 (2.404/2.496) \\
 & MSE & 9.695 (9.560/9.855) & 9.829 (9.722/9.974) & 9.526 (9.167/10.016) \\
 & SMAPE & 31.713 (31.422/31.989) & 31.723 (31.472/31.999) & 31.190 (30.582/32.127) \\
\midrule
\multirow{4}{*}{PATCHTST LPE} & RMSE & 2.984 (2.961/3.021) & 3.001 (2.988/3.012) & 2.978 (2.949/3.001) \\
 & MAE & 2.338 (2.323/2.365) & 2.369 (2.359/2.375) & 2.357 (2.332/2.375) \\
 & MSE & 8.908 (8.768/9.135) & 9.007 (8.927/9.073) & 8.872 (8.704/9.010) \\
 & SMAPE & 30.099 (29.812/30.558) & 30.058 (29.838/30.196) & 29.828 (29.540/30.068) \\
\cmidrule(lr){2-5}
\multirow{4}{*}{PATCHTST L-RoPE LPE} & RMSE & 3.083 (3.028/3.141) & 3.040 (3.017/3.072) & 3.000 (2.986/3.016) \\
 & MAE & 2.437 (2.391/2.488) & 2.410 (2.391/2.436) & 2.376 (2.365/2.391) \\
 & MSE & 9.515 (9.171/9.883) & 9.247 (9.103/9.442) & 9.002 (8.918/9.098) \\
 & SMAPE & 31.365 (30.608/32.149) & 30.585 (30.302/30.972) & 30.162 (30.000/30.336) \\
\bottomrule
\end{tabular}
\caption{Full benchmark results on ETTh1 across token budgets. Each cell shows Mean (95\% BCa confidence interval). For each metric, the best (bold) and second-best (underlined) values per dataset are highlighted.}
\label{tab:appendix_results_ci_etth1}
\end{table*}

\begin{table*}[htbp]
\centering
\small
\begin{tabular}{llccc}
\toprule
\textbf{Model} & \textbf{Metric} & \textbf{45} & \textbf{90} & \textbf{180} \\ \midrule
\multirow{4}{*}{BSAT LPE} & RMSE & 4.518 (4.502/4.548) & 4.728 (4.691/4.791) & 4.840 (4.698/5.021) \\
 & MAE & 3.060 (3.047/3.090) & 3.280 (3.258/3.312) & 3.379 (3.264/3.519) \\
 & MSE & 20.411 (20.272/20.685) & 22.361 (22.006/22.945) & 23.524 (22.116/25.236) \\
 & SMAPE & 151.124 (151.016/151.195) & 150.892 (150.815/150.966) & 150.195 (150.073/150.323) \\
\cmidrule(lr){2-5}
\multirow{4}{*}{BSAT L-RoPE LPE} & RMSE & \underline{4.124 (4.112/4.136)} & 4.423 (4.362/4.481) & 4.351 (4.284/4.428) \\
 & MAE & 2.706 (2.687/2.731) & 2.937 (2.900/2.989) & 2.854 (2.803/2.912) \\
 & MSE & \underline{17.005 (16.905/17.110)} & 19.574 (19.037/20.097) & 18.952 (18.353/19.632) \\
 & SMAPE & 149.473 (149.421/149.528) & 149.357 (149.287/149.432) & 149.359 (149.268/149.490) \\
\midrule
\multirow{4}{*}{UDS LPE} & RMSE & 4.351 (4.315/4.388) & 4.399 (4.230/4.707) & 4.138 (4.042/4.266) \\
 & MAE & 2.735 (2.702/2.778) & 2.817 (2.657/3.132) & \underline{2.623 (2.501/2.750)} \\
 & MSE & 18.940 (18.629/19.251) & 19.550 (17.956/22.612) & 17.173 (16.385/18.272) \\
 & SMAPE & 148.887 (148.840/148.928) & 148.657 (148.492/148.936) & \textbf{148.098 (148.066/148.138)} \\
\cmidrule(lr){2-5}
\multirow{4}{*}{UDS L-RoPE LPE} & RMSE & 4.305 (4.270/4.334) & 4.341 (4.191/4.514) & \textbf{4.038 (3.896/4.382)} \\
 & MAE & 2.672 (2.647/2.700) & 2.693 (2.551/2.846) & \textbf{2.422 (2.286/2.772)} \\
 & MSE & 18.541 (18.239/18.790) & 18.951 (17.615/20.505) & \textbf{16.464 (15.195/19.985)} \\
 & SMAPE & 148.976 (148.927/149.068) & 148.849 (148.678/149.018) & \underline{148.171 (148.019/148.363)} \\
\midrule
\multirow{4}{*}{PATCHTST LPE} & RMSE & 4.778 (4.427/5.379) & 5.229 (4.845/5.714) & 5.210 (4.816/5.740) \\
 & MAE & 3.254 (2.918/3.769) & 3.733 (3.396/4.129) & 3.393 (3.103/3.762) \\
 & MSE & 23.612 (20.033/30.464) & 28.077 (23.957/33.613) & 27.966 (23.804/34.601) \\
 & SMAPE & 149.005 (148.745/149.404) & 149.248 (148.882/149.653) & 149.161 (148.709/149.989) \\
\cmidrule(lr){2-5}
\multirow{4}{*}{PATCHTST L-RoPE LPE} & RMSE & 4.634 (4.379/4.993) & 4.847 (4.489/5.331) & 4.405 (4.127/4.908) \\
 & MAE & 3.097 (2.849/3.408) & 3.127 (2.847/3.621) & 2.748 (2.534/3.107) \\
 & MSE & 21.828 (19.411/25.493) & 24.155 (20.611/29.542) & 19.934 (17.216/25.162) \\
 & SMAPE & 149.020 (148.764/149.294) & 149.290 (148.964/149.737) & 148.397 (148.080/148.884) \\
\bottomrule
\end{tabular}
\caption{Full benchmark results on Alabama PV across token budgets. Each cell shows Mean (95\% BCa confidence interval). For each metric, the best (bold) and second-best (underlined) values per dataset are highlighted.}
\label{tab:appendix_results_ci_al_solar06}
\end{table*}

\begin{table*}[htbp]
\centering
\small
\begin{tabular}{llccc}
\toprule
\textbf{Model} & \textbf{Metric} & \textbf{45} & \textbf{90} & \textbf{180} \\ \midrule
\multirow{4}{*}{BSAT LPE} & RMSE & 11.004 (10.981/11.035) & 10.599 (10.596/10.602) & 10.527 (10.523/10.531) \\
 & MAE & 7.956 (7.936/7.983) & 7.593 (7.590/7.598) & 7.528 (7.524/7.535) \\
 & MSE & 121.094 (120.591/121.814) & 112.331 (112.264/112.398) & 110.812 (110.741/110.896) \\
 & SMAPE & 7.698 (7.678/7.724) & 7.342 (7.339/7.347) & 7.269 (7.266/7.274) \\
\cmidrule(lr){2-5}
\multirow{4}{*}{BSAT L-RoPE LPE} & RMSE & 10.786 (10.769/10.800) & 10.674 (10.616/10.743) & 10.601 (10.391/10.810) \\
 & MAE & 7.828 (7.813/7.844) & 7.748 (7.693/7.816) & 7.658 (7.463/7.849) \\
 & MSE & 116.345 (115.973/116.632) & 113.956 (112.743/115.410) & 112.549 (107.965/116.839) \\
 & SMAPE & 7.563 (7.549/7.579) & 7.492 (7.438/7.556) & 7.401 (7.216/7.581) \\
\midrule
\multirow{4}{*}{UDS LPE} & RMSE & 11.827 (11.777/11.880) & 11.342 (11.259/11.437) & 10.435 (10.280/10.890) \\
 & MAE & 8.596 (8.558/8.641) & 8.248 (8.163/8.331) & 7.496 (7.371/7.890) \\
 & MSE & 139.883 (138.744/141.093) & 128.669 (126.822/130.844) & 109.131 (105.737/119.779) \\
 & SMAPE & 8.274 (8.238/8.313) & 7.931 (7.854/8.007) & 7.227 (7.101/7.620) \\
\cmidrule(lr){2-5}
\multirow{4}{*}{UDS L-RoPE LPE} & RMSE & 11.442 (11.398/11.537) & 11.322 (11.228/11.399) & 10.433 (10.337/10.554) \\
 & MAE & 8.214 (8.179/8.282) & 8.173 (8.095/8.252) & 7.522 (7.447/7.619) \\
 & MSE & 130.942 (129.907/133.075) & 128.220 (126.064/129.939) & 108.889 (106.854/111.498) \\
 & SMAPE & 7.951 (7.920/8.014) & 7.871 (7.799/7.943) & 7.242 (7.166/7.338) \\
\midrule
\multirow{4}{*}{PATCHTST LPE} & RMSE & 9.293 (9.283/9.304) & 9.305 (9.295/9.314) & \textbf{9.290 (9.284/9.296)} \\
 & MAE & 6.594 (6.584/6.605) & 6.605 (6.596/6.613) & \underline{6.583 (6.579/6.590)} \\
 & MSE & 86.357 (86.180/86.565) & 86.577 (86.400/86.749) & \textbf{86.295 (86.195/86.403)} \\
 & SMAPE & 6.361 (6.352/6.371) & 6.371 (6.362/6.379) & \underline{6.353 (6.348/6.358)} \\
\cmidrule(lr){2-5}
\multirow{4}{*}{PATCHTST L-RoPE LPE} & RMSE & 9.337 (9.314/9.373) & 9.291 (9.284/9.301) & \underline{9.290 (9.286/9.298)} \\
 & MAE & 6.635 (6.612/6.674) & 6.590 (6.583/6.600) & \textbf{6.577 (6.573/6.583)} \\
 & MSE & 87.177 (86.747/87.879) & 86.317 (86.185/86.507) & \underline{86.309 (86.219/86.450)} \\
 & SMAPE & 6.402 (6.381/6.439) & 6.358 (6.351/6.368) & \textbf{6.346 (6.341/6.351)} \\
\bottomrule
\end{tabular}
\caption{Full benchmark results on ECL across token budgets. Each cell shows Mean (95\% BCa confidence interval). For each metric, the best (bold) and second-best (underlined) values per dataset are highlighted.}
\label{tab:appendix_results_ci_ecl}
\end{table*}

\subsubsection{Additional Visualizations}
\begin{figure*}[t]
  \centering
  \includegraphics[width=0.95\textwidth]{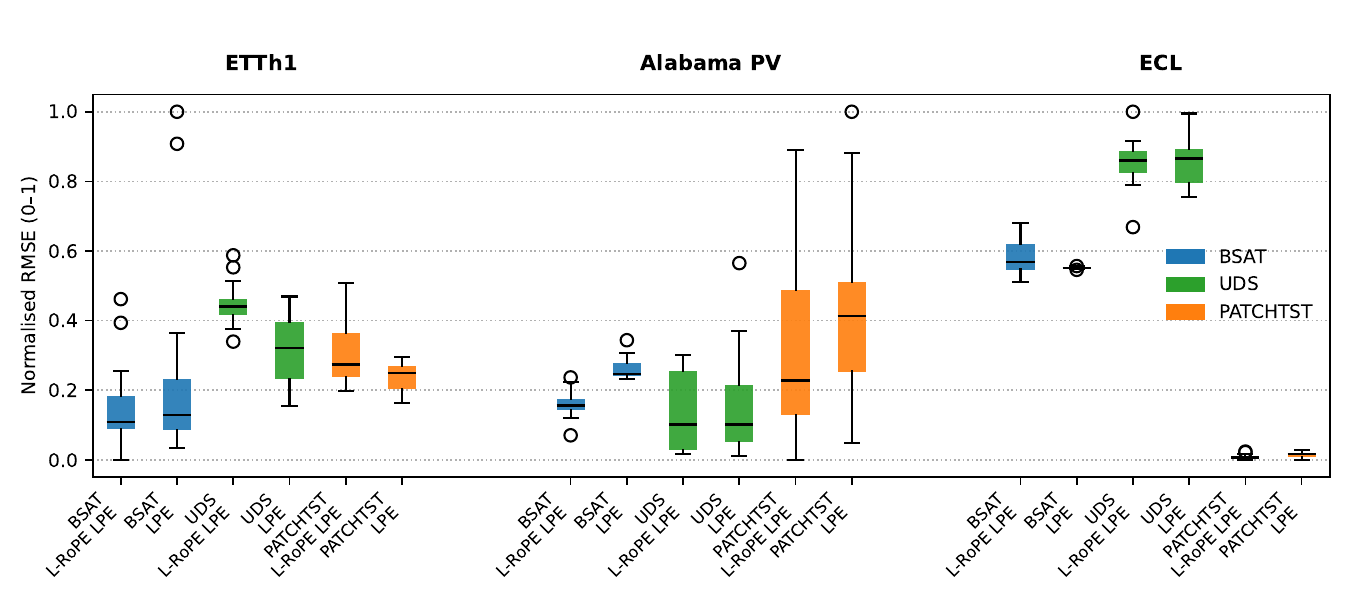}
  \caption{Box-plot for token budget 90 across all datasets, showing the distribution of the top 15 runs for each configuration. RMSE is rescaled to $[0,1]$ inside each dataset~$\times$~budget group.}
  \label{fig:box_90}
\end{figure*}

\begin{figure*}[t]
  \centering
  \includegraphics[width=0.95\textwidth]{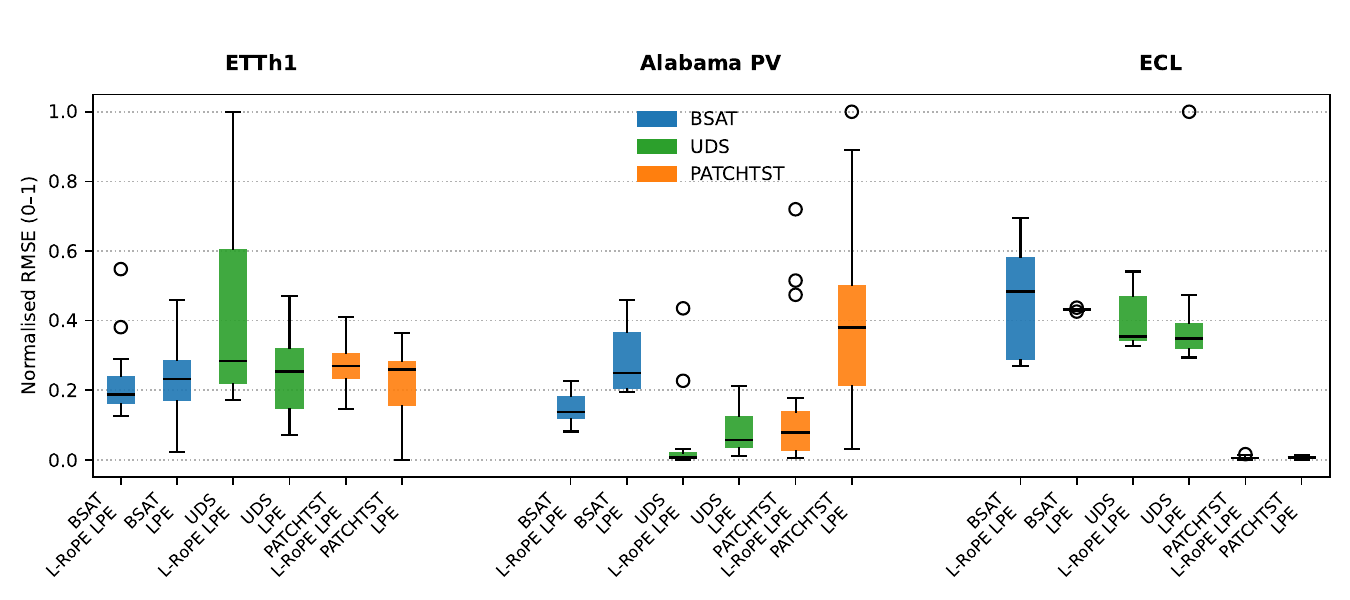}
  \caption{Box-plot for token budget 180 across all datasets, showing the distribution of the top 15 runs for each configuration. RMSE is rescaled to $[0,1]$ inside each dataset~$\times$~budget group.}
  \label{fig:box_180}
\end{figure*}

To visualize performance and reliability, box-plots at token budgets 45 \cref{fig:box_45}, 90 \cref{fig:box_90} and \cref{fig:box_180} are provided. To compare all datasets fairly, results are min--max scaled within each dataset $\times$ token-budget group: $x' = (x - \min)/(\max - \min)$.
Across datasets and token budgets BSAT shows strong minimum and median results, as well as high stability.


\clearpage
\small{
\bibliography{bibliography}
}
\clearpage
\makeatletter
\@ifundefined{isChecklistMainFile}{
  % We are compiling a standalone document
  \newif\ifreproStandalone
  \reproStandalonetrue
}{
  % We are being \input into the main paper
  \newif\ifreproStandalone
  \reproStandalonefalse
}
\makeatother

\ifreproStandalone
\documentclass[letterpaper]{article}
\usepackage[submission]{aaai2026}
\setlength{\pdfpagewidth}{8.5in}
\setlength{\pdfpageheight}{11in}
\usepackage{times}
\usepackage{helvet}
\usepackage{courier}
\usepackage{xcolor}
\frenchspacing

\begin{document}
\fi
\setlength{\leftmargini}{20pt}
\makeatletter\def\@listi{\leftmargin\leftmargini \topsep .5em \parsep .5em \itemsep .5em}
\def\@listii{\leftmargin\leftmarginii \labelwidth\leftmarginii \advance\labelwidth-\labelsep \topsep .4em \parsep .4em \itemsep .4em}
\def\@listiii{\leftmargin\leftmarginiii \labelwidth\leftmarginiii \advance\labelwidth-\labelsep \topsep .4em \parsep .4em \itemsep .4em}\makeatother

\setcounter{secnumdepth}{0}
\renewcommand\thesubsection{\arabic{subsection}}
\renewcommand\labelenumi{\thesubsection.\arabic{enumi}}

\newcounter{checksubsection}
\newcounter{checkitem}[checksubsection]

\newcommand{\checksubsection}[1]{%
  \refstepcounter{checksubsection}%
  \paragraph{\arabic{checksubsection}. #1}%
  \setcounter{checkitem}{0}%
}

\newcommand{\checkitem}{%
  \refstepcounter{checkitem}%
  \item[\arabic{checksubsection}.\arabic{checkitem}.]%
}
\newcommand{\question}[2]{\normalcolor\checkitem #1 #2 \color{blue}}
\newcommand{\ifyespoints}[1]{\makebox[0pt][l]{\hspace{-15pt}\normalcolor #1}}

\section*{Reproducibility Checklist}

\vspace{1em}
\hrule
\vspace{1em}

\textbf{Instructions for Authors:}

This document outlines key aspects for assessing reproducibility. Please provide your input by editing this \texttt{.tex} file directly.

For each question (that applies), replace the ``Type your response here'' text with your answer.

\vspace{1em}
\noindent
\textbf{Example:} If a question appears as
\begin{center}
\noindent
\begin{minipage}{.9\linewidth}
\ttfamily\raggedright
\string\question \{Proofs of all novel claims are included\} \{(yes/partial/no)\} \\
Type your response here
\end{minipage}
\end{center}
you would change it to:
\begin{center}
\noindent
\begin{minipage}{.9\linewidth}
\ttfamily\raggedright
\string\question \{Proofs of all novel claims are included\} \{(yes/partial/no)\} \\
yes
\end{minipage}
\end{center}
Please make sure to:
\begin{itemize}\setlength{\itemsep}{.1em}
\item Replace ONLY the ``Type your response here'' text and nothing else.
\item Use one of the options listed for that question (e.g., \textbf{yes}, \textbf{no}, \textbf{partial}, or \textbf{NA}).
\item \textbf{Not} modify any other part of the \texttt{\string\question} command or any other lines in this document.\\
\end{itemize}

You can \texttt{\string\input} this .tex file right before \texttt{\string\end\{document\}} of your main file or compile it as a stand-alone document. Check the instructions on your conference's website to see if you will be asked to provide this checklist with your paper or separately.

\vspace{1em}
\hrule
\vspace{1em}

% The questions start here

\checksubsection{General Paper Structure}
\begin{itemize}

\question{Includes a conceptual outline and/or pseudocode description of AI methods introduced}{(yes/partial/no/NA)}
yes

\question{Clearly delineates statements that are opinions, hypothesis, and speculation from objective facts and results}{(yes/no)}
yes

\question{Provides well-marked pedagogical references for less-familiar readers to gain background necessary to replicate the paper}{(yes/no)}
yes

\end{itemize}
\checksubsection{Theoretical Contributions}
\begin{itemize}

\question{Does this paper make theoretical contributions?}{(yes/no)}
no

	\ifyespoints{\vspace{1.2em}If yes, please address the following points:}
        \begin{itemize}
	
	\question{All assumptions and restrictions are stated clearly and formally}{(yes/partial/no)}
	Type your response here

	\question{All novel claims are stated formally (e.g., in theorem statements)}{(yes/partial/no)}
	Type your response here

	\question{Proofs of all novel claims are included}{(yes/partial/no)}
	Type your response here

	\question{Proof sketches or intuitions are given for complex and/or novel results}{(yes/partial/no)}
	Type your response here

	\question{Appropriate citations to theoretical tools used are given}{(yes/partial/no)}
	Type your response here

	\question{All theoretical claims are demonstrated empirically to hold}{(yes/partial/no/NA)}
	Type your response here

	\question{All experimental code used to eliminate or disprove claims is included}{(yes/no/NA)}
	Type your response here
	
	\end{itemize}
\end{itemize}

\checksubsection{Dataset Usage}
\begin{itemize}

\question{Does this paper rely on one or more datasets?}{(yes/no)}
yes

\ifyespoints{If yes, please address the following points:}
\begin{itemize}

	\question{A motivation is given for why the experiments are conducted on the selected datasets}{(yes/partial/no/NA)}
yes

	\question{All novel datasets introduced in this paper are included in a data appendix}{(yes/partial/no/NA)}
NA

	\question{All novel datasets introduced in this paper will be made publicly available upon publication of the paper with a license that allows free usage for research purposes}{(yes/partial/no/NA)}
NA

	\question{All datasets drawn from the existing literature (potentially including authors' own previously published work) are accompanied by appropriate citations}{(yes/no/NA)}
yes

	\question{All datasets drawn from the existing literature (potentially including authors' own previously published work) are publicly available}{(yes/partial/no/NA)}
yes

	\question{All datasets that are not publicly available are described in detail, with explanation why publicly available alternatives are not scientifically satisficing}{(yes/partial/no/NA)}
NA

\end{itemize}
\end{itemize}

\checksubsection{Computational Experiments}
\begin{itemize}

\question{Does this paper include computational experiments?}{(yes/no)}
yes

\ifyespoints{If yes, please address the following points:}
\begin{itemize}

	\question{This paper states the number and range of values tried per (hyper-) parameter during development of the paper, along with the criterion used for selecting the final parameter setting}{(yes/partial/no/NA)}
yes

	\question{Any code required for pre-processing data is included in the appendix}{(yes/partial/no)}
partial

	\question{All source code required for conducting and analyzing the experiments is included in a code appendix}{(yes/partial/no)}
partial
	\question{All source code required for conducting and analyzing the experiments will be made publicly available upon publication of the paper with a license that allows free usage for research purposes}{(yes/partial/no)}
yes

	\question{All source code implementing new methods have comments detailing the implementation, with references to the paper where each step comes from}{(yes/partial/no)}
yes

	\question{If an algorithm depends on randomness, then the method used for setting seeds is described in a way sufficient to allow replication of results}{(yes/partial/no/NA)}
yes

	\question{This paper specifies the computing infrastructure used for running experiments (hardware and software), including GPU/CPU models; amount of memory; operating system; names and versions of relevant software libraries and frameworks}{(yes/partial/no)}
yes

	\question{This paper formally describes evaluation metrics used and explains the motivation for choosing these metrics}{(yes/partial/no)}
yes

	\question{This paper states the number of algorithm runs used to compute each reported result}{(yes/no)}
yes

	\question{Analysis of experiments goes beyond single-dimensional summaries of performance (e.g., average; median) to include measures of variation, confidence, or other distributional information}{(yes/no)}
yes

	\question{The significance of any improvement or decrease in performance is judged using appropriate statistical tests (e.g., Wilcoxon signed-rank)}{(yes/partial/no)}
no

	\question{This paper lists all final (hyper-)parameters used for each model/algorithm in the paper’s experiments}{(yes/partial/no/NA)}
partial

\end{itemize}
\end{itemize}
\ifreproStandalone
\end{document}
\fi
\end{document}